\documentclass[journal]{IEEEtran}  

\usepackage{amsmath} 
\usepackage{amsfonts}
\usepackage{tabularx}
\usepackage{graphicx}
\usepackage{subcaption}
\usepackage{multirow}
\usepackage{xcolor}
\title{\LARGE \bf
Deep Learning based Pedestrian Inertial Navigation: \\  Methods, Dataset and On-Device Inference
}

\begin{document}

\author{Changhao Chen, Peijun Zhao, Chris Xiaoxuan Lu, Wei Wang, Andrew Markham, Niki Trigoni
\thanks{*The authors are with the Department of Computer Science, University of Oxford, Oxford OX1 3QD, U.K.
        (Email: {\tt\small changhao.chen@cs.ox.ac.uk; peijun.zhao@cs.ox.ac.uk; xiaoxuan.lu@cs.ox.ac.uk; wei.wang@cs.ox.ac.uk; andrew.markham@cs.ox.ac.uk; niki.trigoni@cs.ox.ac.uk})(The first two authors contributed equally to this work.)(Corresponding author: Chris Xiaoxuan Lu)}%
\thanks{*This work was supported by EPSRC Program Grant Mobile Robotics: Enabling a Pervasive Technology of the Future (GoW EP/M019918/1) and the National Institute of Standards and Technology (NIST) via the grant Pervasive, Accurate, and Reliable Location-based Services for Emergency Responders (Federal Grant: 70NANB17H185).}

}

\markboth{IEEE INTERNET OF THINGS JOURNAL,~Vol.~X, No.~X, X~2019}%
{Shell \MakeLowercase{\textit{et al.}}: Bare Demo of IEEEtran.cls for IEEE Journals}

\maketitle


\begin{abstract}

Modern inertial measurements units (IMUs) are small, cheap, energy efficient, and widely employed in smart devices and mobile robots. Exploiting inertial data for accurate and reliable pedestrian navigation supports is a key component for emerging Internet-of-Things applications and services. 
Recently, there has been a growing interest in applying deep neural networks (DNNs) to motion sensing and location estimation.
However, the lack of sufficient labelled data for training and evaluating architecture benchmarks has limited the adoption of DNNs in IMU-based tasks. 
In this paper, we present and release the \textit{Oxford Inertial Odometry Dataset (OxIOD)}, a first-of-its-kind public dataset for deep learning based inertial navigation research, with fine-grained ground-truth on all sequences. Furthermore, to enable more efficient inference at the edge, we propose a novel lightweight framework to learn and reconstruct pedestrian trajectories from raw IMU data. Extensive experiments show the effectiveness of our dataset and methods in achieving accurate data-driven pedestrian inertial navigation on resource-constrained devices.

\end{abstract}



\section{INTRODUCTION}

Modern micro-electro-mechanical (MEMS) inertial measurements units (IMUs) are small (a few mm$^2$), cheap (several dollars a piece), energy efficient and pervasive. As a low-cost yet powerful sensing modality, they have received a large amount of research effort and deeply weave into a wide range of applications. For instance, today's smart phones come with embedded IMUs while users can use them for different location-based services, e.g. indoor navigation, localisation and outdoor trajectory analysis \cite{Harle2013}. Moreover, emerging cyber gadgets, such as wristbands and VR/AR headsets, also actively utilise IMUs to provide continuous health monitoring \cite{Gowda2017}, accurate activity tagging \cite{Bianchi2019} and immersive gaming experiences \cite{Marchand2016}. On the side of robots and autonomous systems, IMUs are a long-standing sensing solution to navigation and grasping tasks \cite{Leutenegger2015}. 

The proliferation of IMUs in the aforementioned applications depends on a method called inertial navigation (aka. intertial odometry). Inertial navigation produces orientation and position of users based on the rotation and acceleration measurements of IMU sensors. Such a method is a pillar to motion sensing, acting as a key enabler for many location based services. Compared with GPS, vision, radio or other navigation techniques \cite{Kuutti2018}, 
 the inertial solution relies only on self-contained sensor, requires few physical infrastructure, and is insensitive to environmental dynamics. This unique property, coupled with the proliferation of IMUs in smart devices, allows the flexibility and reliability to deploy the location service easily to IoT applications. 
 Meanwhile, compared with high-dimensional visual data, IMU data are 6-dim time series that can be processed in real time even on resource-constrained device. As such, user's location/motion privacy is thus better protected off the cloud.

	\begin{figure}
    	\centering
        \includegraphics[width=0.48\textwidth]{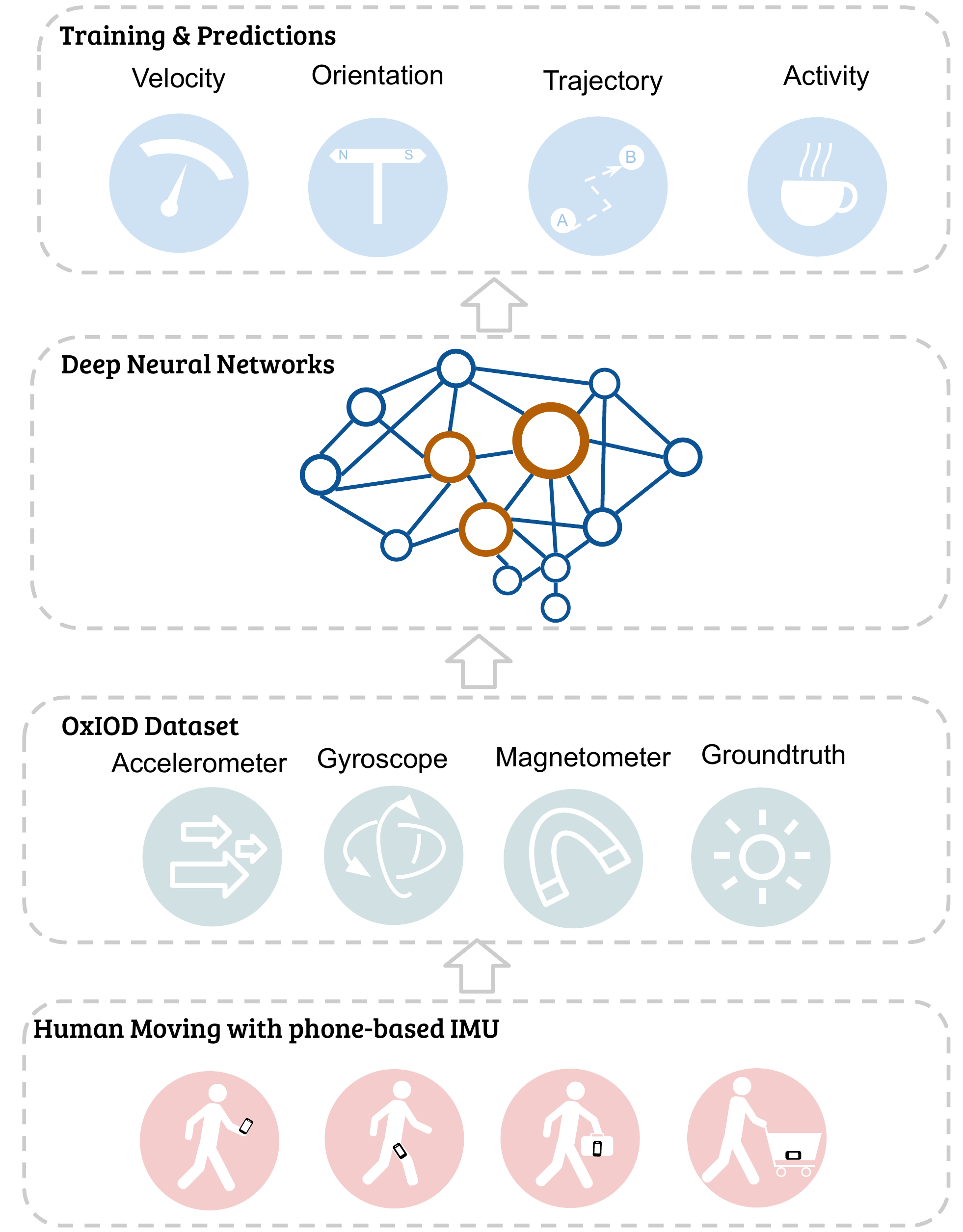}
        \caption{\label{fig:comparison} Deep learning based inertial odometry models can learn and predict human motion from raw inertial data.
        }
    \end{figure} 
    
\begin{table*}[ht]
\caption{Comparison of datasets with IMU and ground truth}
\label{dataset_compare}
\begin{center}
\begin{tabular}{c c c c c c c c c}
\hline
Dataset & Year & Environment & Attachment & IMU Type & Sample Rate & Groundtruth &  Accuracy & Data Size\\
\hline
KITTI Odometry & 2013 & Outdoors & Car & OXTS RT3003 & 10 Hz & GPS/IMU & 10 cm & 22 seqs, 39.2 km\\
EuRoC MAV & 2016 & Indoors & MAV & ADIS 16488 & 200 Hz & Motion Capture & 1 mm & 11 seqs, 0.9 km\\
Oxford RobotCar & 2016 & Outdoors & Car & NovAte SPAN & 50 Hz & GPS/IMU & Unknown & 1010.46 km\\
TUM VI & 2018 & In/Outdoors & Human & BMI 160 & 200 Hz & Motion Capture & 1 mm & 28 seqs, 20 km\\
ADVIO & 2018 & In/Outdoors & Human & InvenSense 20600 & 100 Hz & Other Algorithms & Unknown & 23 seqs, 4.5 km\\
\hline
\textbf{OxIOD (Ours)} & 2018 & Indoors & Human & InvenSense 20600 
& 100 Hz & Motion Capture & 0.5 mm & \textbf{158 seqs}, \textbf{42.587 km}\\
\hline
\end{tabular}
\end{center}
\end{table*}

To achieve long term inertial navigation, a major limitation is the unbounded system error growth, caused by various sensor errors and biases due to the use of low-cost IMUs \cite{Naser2008}. 
Most previous work has exploited human motion context information to constrain the error drifts of the inertial systems.
One solution is to attach the IMU on a user's foot to take advantage of zero-velocity update (ZUPT) for compensating the system drift \cite{Nilsson2012}. 
Pedestrian dead reckoning systems (PDRs) \cite{Harle2013} have been proposed to estimate trajectories by detecting steps and estimating heading. 
However, these handcrafted algorithms are hard to apply in everyday usage due to the unrealistic assumptions of human motion: ZUPT requires the inertial sensor to be firmly fixed on a user's foot, preventing this solution from being used on consumer devices; PDRs are based on personal walking models, and constrained only to work under periodic pedestrian motion. 

Recently, deep learning based inertial navigation models, e.g., IONet \cite{Chen2018}, are proved to be capable of estimating motion and generating trajectories directly from raw inertial data without any handcrafted engineering. Other data-driven methods \cite{Yan2018,Cortes2018} learn to predict velocities in order to constrain system error drift, and achieve competitive performance. These learning based models have been shown to outperform previous model-based approaches in terms of accuracy and robustness \cite{Chen2018,Yan2018,Cortes2018}. There is a growing interest in applying deep neural networks to learn motion from time-series data, due to its potential for model-free generalisation.

However, to develop data-driven approaches, we are confronted with the following three main challenges: 
1) A significant amount of sensor data with highly precise labels, i.e. the ground-truth values of location, velocity and orientation are required for training, validating and testing deep neural network models. Existing datasets \cite{Geiger2013,Maddern2016,Schubert2018,Cortes2018-1} are not suitable for training DNN models for human tracking, as the sensor data are collected either from vehicles e.g. cars, or fixed in specific position, which can not reflect the IMU motion in everyday usage e.g. as would be sensed by a smartphone. 
2) Few works have considered the efficiency of deep neural network models for inertial odometry when deployed on low-end devices. It is important for machine learning models to run at the edge close to where the sensor data are collected, as this will improve the reliability and latency of the inference, and protect the users' privacy \cite{Samie2019}, particularly in IoT applications.
3) There is a lack of common evaluation benchmarks, whether for conventional PDRs or learning based models, which precludes a fair and objective comparison of different techniques. 

In this paper, we present and release the Oxford Inertial Odometry Dataset (OxIOD), with a large amount of pedestrian, multi-attachment sensor data (158 sequences, totalling more than 42 km in distance), and high-precise labels, much larger than prior inertial navigation datasets. 
In order to capture human motion that accurately reflects everyday usage, the data were collected with a high degree of diversity, across different attachments, motion modes, users, types of device and places. 
As illustrated in Figure \ref{fig:comparison}, our proposed dataset is able to be used to train robust and accurate deep learning models for inertial navigation, and we evaluate both the classical algorithms (PDRs) and data-driven models on OxIOD as a common benchmark. 
To enhance the online efficiency of DNN models on mobile devices, we propose Light Inertial Odometry Neural Networks (L-IONet), a lightweight deep neural network framework to learn inertial navigation from raw data without any handcrafted engineering, which is much more efficient at training and inference than previously proposed models using LSTM (Long Short-Term Memory neural network). 
Extensive experiments were conducted to evaluate the proposed model and existing methods for a systematic study into the performance of the data-driven inertial odometry models in real-world applications and inference at the edge. 

In summary, we have three main contributions:

\begin{itemize}
    \item We present OxIOD\footnote{The OxIOD Dataset is available at: http://deepio.cs.ox.ac.uk}, a first-of-its-kind dataset for pedestrian inertial navigation research, to both boost the adoption of data-driven methods and provide a common benchmark for the task of pedestrian inertial navigation. 
    \item We propose L-IONet, a lightweight deep neural network framework to efficiently learn and infer inertial odometry from raw IMU data.
    \item We conduct a systematic research into the computational and runtime efficiency of deep neural network models deployed on low-end mobile devices.
\end{itemize}

The rest of the paper is organised as follows. Section 2 surveys the related work on the existing datasets and models. Section 3 introduces the Oxford Inertial Odometry Dataset. In Section 4, we present a novel lightweight learning based inertial odometry model. Section 5 provides comprehensive evaluations and results.

\section{RELATED WORK}

\subsection{Inertial Navigation Datasets}

Table \ref{dataset_compare} shows representative datasets that include inertial data for the purpose of navigation and localisation.
In KITTI \cite{Geiger2013}, Oxford RobotCar \cite{Maddern2016} and EuRoC MAV datasets \cite{Burri2016}, the sensors are rigidly fixed to the chassis of a car, which is suitable for studying vehicle movements, but not directly useful for studying human movement. The TUM VI dataset \cite{Schubert2018} was collected to evaluate visual-inertial odometry (VIO), with a pedestrian holding the device in front of them. The ground truth in TUM VI is provided at the beginning and ending of the sequences, while during most of the trajectories there is no ground truth. Similarly, in ADVIO \cite{Cortes2018-1}, the dataset is rather short (4.5 km) and only offers pseudo ground truth generated by a handcrafted inertial odometry algorithm. 

There are several datasets focusing on human gait and activities, which are somewhat similar to our dataset, but do not concentrate on localisation. Some of these datasets measure human activities, such as USC-HAD \cite{Zhang2012}, CMU-MMAC \cite{DeLaTorre2008}, and OPPORTUNITY \cite{Chavarriaga2013}. Though these datasets have inertial data with accurate poses as ground truth, they cannot be used to train and test odometry/localisation, since the participants did not move much during the experiments.
Some other datasets, such as MAREA \cite{Siddhartha2017}, focus on human gait recognition and collected inertial data while carriers were walking or running. However, these datasets lack solid ground truth and thus limit their usage in training and testing odometry models.

As we can see from Table \ref{dataset_compare}, our OxIOD dataset has a huge amount of data from 158 sequences, leading to a total distance of 42.587km. The data size of OxIOD is larger than most other inertial navigation datasets, and hence is suitable for deep neural network methods, which require large amounts of data and high accuracy labels. It should be noted that the total length of the dataset even exceeds those collected by vehicles. 
Meanwhile, our dataset can better represent human motion in everyday conditions and thus has a greater diversity.

\subsection{Inertial Navigation Using Low-cost IMUs}

Due to high sensor noise and bias, it is impossible to use conventional Strapdown Inertial Navigation Systems (SINS), which directly integrate inertial measurements into orientation, velocity and location, on low-cost MEMS IMU platforms. 
To realise purely inertial pedestrian navigation, most of existing methods exploit domain specific knowledge to constrain the error drift of inertial systems. One solution is to attach inertial sensor onto users' foot to take advantage of stationary phases during human walking to perform the zero-velocity update (ZUPT). The ZUPT based method can be further enhanced by known velocity update and double-foot position calibration \cite{Diliang2019}.  
Another solution is Pedestrian Dead Reckoning (PDR), very common in phone based pedestrian navigation. Under the assumption that users exhibit periodic motion, PDRs update locations by counting users' steps and estimating their stride length and heading \cite{Harle2013}. 
Recent research focuses on fusing other sensor modalities with the PDR models to further improve the robustness and accuracy, such as wireless signals \cite{Zhuang2018}, magnetic fields \cite{Li2019,Wang2016a} or UWB \cite{LiuYDJLT17}. 

Recent emerging data-driven solutions are capable of learning a more general motion model from a large amount of inertial data without hand-engineering effort.
A good example is IONet \cite{Chen2018}, which proposes to formulate inertial odometry as a sequential learning problem and constructs a deep recurrent neural network (RNN) framework to reconstruct trajectories directly from raw inertial data, outperforming traditional model-based methods. Other methods learn to recover latent velocities \cite{Yan2018} \cite{Cortes2018}, or detect more accurate zero-velocity phase, in order to compensate the errors of inertial systems \cite{Wagstaff2018}. 
However, few of the previous works considers the inference efficiency of deep learning approaches when deployed on low-end devices. 

    \begin{figure}
    	\centering
        \includegraphics[width=0.5\textwidth]{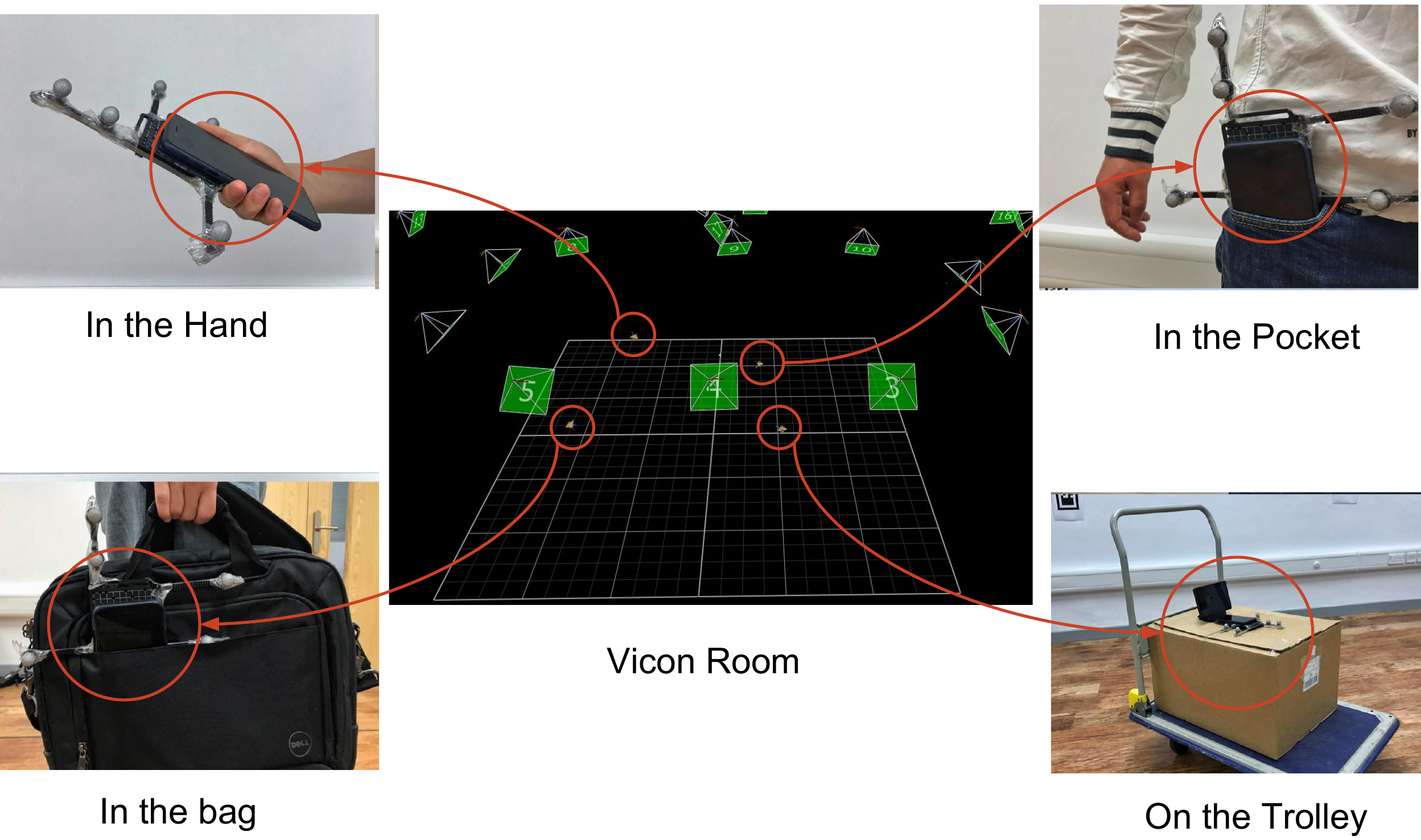}
        \caption{\label{fig:attachments} Inertial data are collected from a smartphone in four different attachments: handheld (left above), pocket (right above), handbag (left below), trolley (right below). The high-precise motion labels are provided by the Vicon System.}
    \end{figure}

\section{Oxford Inertial Odometry Dataset}
This section introduces the Oxford Inertial Odometry Dataset (OxIOD), a data collection of inertial measurements for training and evaluating deep learning based inertial odometry models. To reflect sensor readings under everyday usage, the data were collected with IMUs with various attachments (handheld, in the pocket, in the handbag and on a trolley/stroller), motion modes (halting, walking slowly, walking normally, and running), 
four types of off-the-shelf consumer phones and five different users, as illustrated in Table \ref{tb:dataset}. 
Our dataset has 158 sequences, and the total walking distance and recording time are 42.5 km, and 14.72 h (53022 seconds).

\begin{table}[ht]
\caption{Oxford Inertial Odometry Dataset}
\label{tb:dataset}
\begin{center}
\begin{tabular}{c c c c c}
\hline
~ & Type & Seqs & Time (s) & Distance (km) \\
\hline
\multirow{4}*{Attachments} & Handheld & 24 & 8821 & 7.193\\
~ & Pocket & 11 & 5622 & 4.231\\
{\tiny (iPhone 7P/User 1)} & Handbag & 8 & 4100 & 3.431\\
{\tiny (Normally Walking)} & Trolley & 13 & 4262 & 2.685\\
\hline
\multirow{3}*{Motions} & Slowly Walking & 8 & 4150 & 2.421\\
~ & Normally Walking & -  & - & - \\
~ & Running & 7 & 3732 & 4.356\\
\hline
\multirow{3}*{Devices} & iPhone 7P & - & - & - \\ 
~ & iPhone 6 & 9 & 1592 & 1.381\\

~ & iPhone 5 & 9 & 1531 & 1.217\\
~ & Nexus 5 & 8 & 4021 & 2.752\\
\hline
\multirow{4}*{Users} & User 1 & - & - & - \\ 
~ & User 2 & 9 & 2928 & 2.422\\
~ & User 3 & 7 & 2100 & 1.743\\
~ & User 4 & 9 & 3118 & 2.812\\
~ & User 5 & 10 & 2884 & 2.488\\
\hline
\multirow{2}*{Large Scale} & floor 1 & 10 & 1579 & 1.412\\
~ & floor 2 & 16 & 2582 & 2.053\\
\hline
Total &  & 158 & 53022 & 42.587\\
\hline
\end{tabular}
\end{center}
\end{table}

\subsection{Sensor Setup}
The data were collected by the on-board sensors of consumer phones, recording accelerations and angular rates from 6-axis IMUs, and magnetic fields from 3-axis magnetometers. The sensor types of IMUs and magnetometers employed in our adopted mobile phones are listed in Table \ref{sensor_list}.
A Vicon motion capture system \cite{Vicon2017} was deployed to produce high-precise groundtruth values of the object motion, i.e. its orientation, velocity and position. The large-scale collection was conducted on two office floors, where we used a Google Tango Tablet \cite{Tango} as pseudo groundtruth.

	\begin{figure*}
    	\centering
        \includegraphics[width=0.9\textwidth]{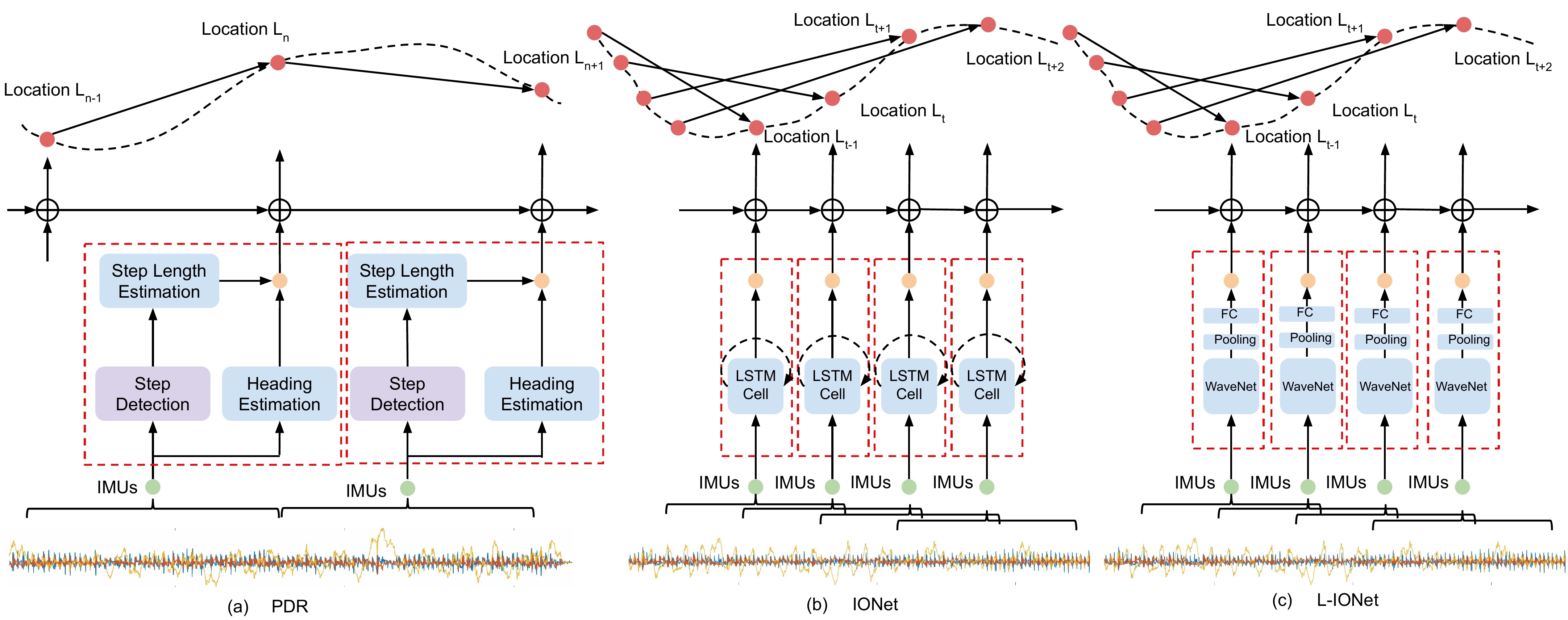}
        \caption{\label{fig:framework} The framework illustration of three inertial navigation models: (a) Pedestrian Dead Reckoning (b) Inertial Odometry Neural Network (IONet) (c) Lightweight Inertial Odometry Neural Network (L-IONet).}
    \end{figure*} 

    \begin{table}[ht]
    \caption{Sensors}
    \label{sensor_list}
        \begin{center}
            \begin{tabular}{c c c}
            \hline
            Mobile Phone & IMU & Magnetometer\\
            \hline
            iPhone 7 Plus & InvenSense ICM-20600 & Alps HSCDTD004A\\
            iPhone 6 & InvenSense MP67B & AKM 8963\\
            iPhone 5 & STL3G4200DH &  AKM 8963\\
            Nexus 5 & InvenSense MPU-6515 & Asahi Kasei AK8963 \\
            \hline
        \end{tabular}
    \end{center}
    \end{table}

\textbf{IMU}:
The majority of data were collected with an iPhone 7 Plus device. 
The IMU inside iPhone 7 Plus is InvenSense ICM-20600, a 6-axis motion tracking sensor. It combines a 3-axis gyroscope and a 3-axis accelerometer. 16-bit ADCs are integrated in both gyroscope and accelerometer. The sensitivity error of the gyroscope is $1 \%$, while the noise is $4  \text{mdps}/\sqrt{\text{Hz}}$. The accelerometer noise is 100 $\mu g/\sqrt{\text{Hz}}$.

\textbf{Magnetometer}
The Alps HSCDTD004A embedded in iPhone 7 Plus is a 3-axis geomagnetic sensor, which is mainly used for electronic compass. It has a measurement range of $\pm 1.2 \text{mT}$ and an output resolution of 0.3 $\mu \text{T/LSB}$.

\textbf{Vicon System}
We deployed 10 Bonita B10 cameras in the Vicon Motion Tracker system \cite{Vicon2017}, encircling an area where we conducted data collection experiments. Each Bonita B10 camera has a frame rate of 250 fps, and resolution of 1 megapixel (1024*1024). The lens operating range of Bonita B10 can be up to 13 m. These features enable us to capture motion data with a precision down to 0.5 mm, making the ground truth very accurate and reliable. The software used in the Vicon system is \textit{Vicon Tracker 2.2}. We connected Vicon Tracker to Robot Operating System (ROS) with \textit{vicon\_bridge}, and recorded the data stream with rostopic. The map size of our experimental setup in Vicon Room is $5 m \times 8 m$.

\textbf{Time Synchronisation}
The IMU and magnetometer are integrated in the mobile phone, sharing the same time stamp. 
Vicon data recorded with ROS is saved with UTC timestamp. Before each experiment, we synchronised the time of iPhone 7 Plus and ROS with UTC, and thus all time stamps recorded along with sensor data will be synchronised.

\subsection{Data Collection}
\textbf{Attachments}
The phone based IMUs will experience distinct motions when attached in different places. In the context of pedestrian navigation, a natural use of mobile phone leads to an unconstrained placement of inertial sensors, and therefore we selected four common situations to study, i.e. the device is in the hand, in the pocket, in the handbag or on the trolley. In our data collection, a pedestrian (named \textit{User 1}) walked naturally inside the Vicon room, carrying a phone in four attachments. Figure \ref{fig:attachments} shows in which way the devices were held during the experiments. 

\textbf{Motion Modes}
Humans move in different motion modes in their everyday activities. We selected and collected data from four typical motion models: halting, walking slowly, walking normally and running. The experiments with different motion modes were performed by \textit{User 1} with iPhone 7Plus in hand, to reduce the influences from user walking habits or sensor properties. The velocities of participants are around 0.5 m/s, 1 m/s, and 1.5 m/s during slow walking, normal walking and running. Our experiments indicate that the user speeding can be directly recovered from raw inertial data via deep neural networks, even under a mixed of motion modes.

\textbf{Devices and Users}
Both sensors properties and the walking habits of users throw impacts on the performance of inertial navigation systems. In order to ensure inertial odometry invariant across devices and users, we collected data from several types of devices and different users. 
Four off-the-shelf smartphone were chosen as experimental devices: iPhone 7 Plus, iPhone 6, iPhone 5, and Nexus 5, listed in Table \ref{sensor_list}. Five participants were recruited to perform experiments with phone in the hand, pocket and handbag respectively. 
The mixed data from various devices and users can also be applied in the identification of devices and users.

\textbf{Large-scale localisation}
Besides the extensive data collection inside the VICON Room, we also conducted large-scale tracking in two environments. Without the help of Vicon system, Google Tango device was chosen to provide pseudo ground truth. Participant was instructed to walk freely in an office building on two separate floors (about 1650 $m^2$ and 2475 $m^2$). The smartphones were placed in the hand, pocket and handbag respectively, while the Tango device was attached on the chest of the participant to capture precise trajectories. Figure \ref{fig:floor1} and Figure \ref{fig:floor2} illustrate the floor maps and pseudo ground truth trajectories captured by Google Tango. 

\section{Inertial Navigation Models}
In this section, we selected and introduced two typical inertial navigation models as our baselines: one is a model-based method, Pedestrian Dead Reckoning (PDR) \cite{Harle2013,Xiao2014}, the other one is a deep learning based method, Inertial Odometry Neural Networks (IONet) \cite{Chen2018}. PDR detects steps, estimates step length and heading and updates locations per step, mitigating exponential increasing drifts of SINS algorithm into linear increasing drifts. IONet is able to learn self-motion directly from raw data above large dataset, and solve more general motion, with advantages of extracting high-level motion representation without hand-engineering. 
A novel lightweight DNN framework, the Light Inertial Odometry Neural Networks (L-IONet), is proposed to enable more accurate and efficient inference for inertial navigation from low-cost IMU data.
Figure \ref{fig:framework} illustrates the frameworks of the PDR, IONet, and L-IONet model.

\subsection{Pedestrian Dead Reckoning}
Pedestrian Dead Reckonings (PDRs) output pairs of [step length, step heading] to construct 2D trajectories on a plane. Instead of naively double integrating inertial measurements, PDR algorithms detect steps and estimate step length from a duration of classified inertial data using human walking model.
We implemented a basic PDR algorithm to quantitatively evaluate its performance on the OxIOD dataset. Aided by the common benchmark, extensions are easy to add on this basic PDR to show the effectiveness of each module.

The PDR models mainly consist of four parts: step detection, step length estimation, heading estimation and location update. 
In our PDR model, the step detection thresholds the mean and variance of accelerations, which further classifies the sensory reading into separate independent strides.
A dynamical step length estimation module uses the Weinberg's empirical equation \cite{Weinberg2002} to produce the location displacement $\Delta l$ during a pedestrian stride. 
Gyroscope signals are integrated into the orientation of inertial sensor, but only the yaw angle is kept as pedestrian heading $\psi$.
The current location $(L^x_{k}, L^y_{k})$ can be updated with the previous location $(L^x_{k-1}, L^y_{k-1})$ via: 
    \begin{equation}
     	\label{eq: location}
    	\left\{
    	\begin{aligned}
    		L^x_{k}=L^x_{k-1}+\Delta l \cos(\psi_{k}) \\
        	L^y_{k}=L^y_{k-1}+\Delta l \sin(\psi_{k}),
        \end{aligned}
       \right.
    \end{equation}
where $\Delta l$ and $\psi_{k}$ are the step length and heading at $k$ th step.  

In real-world practice, it is not always easy for the hand-built algorithms to classify inertial data correctly only based on data patterns. Even if the step detection and classification is accurate, the empirical human walking model to estimate step length is highly correlated with user's walking habits and body properties, causing unavoidable accumulative errors during long-term operating. 

\subsection{Inertial Odometry Neural Networks}
Inertial Odometry Neural Networks (IONet) \cite{Chen2018} are able to learn user's ego-motion directly from raw inertial data and solve more general motions. 
For example, tracking a trolley or other wheeled configurations is quite challenging for PDR models, due to the fact that no walking step or periodicity patterns can be detected in this case. 
In contrast, IONet can regress the location transformation (the average speed) during any fixed window of time, without the explicit components of step detection and step length estimation as in PDRs.

We implemented and trained the IONet model on the OxIOD dataset, to show the effectiveness of OxIOD for data-driven approaches.
The continuous inertial readings are segmented into independent sequences of $n$ frames IMU data  $\{(\mathbf{a}_i, \mathbf{w}_i)\}_{i=1}^{n}$, consisting of 3-dimensional accelerations $\mathbf{a}_i \in \mathbb{R}^3$ and 3-dimensional angular rates $\mathbf{w}_i \in \mathbb{R}^3$ at the time step $i$. The 6-dimensional inertial data are preprocessed to normalise the accelerations and angular rates into a same scale. 
The generated sequences are further feed into the recurrent neural networks (RNNs), e.g. LSTMs to extract effective features for motion estimation. Specifically, Figure \ref{fig:wavenet} (a) illustrates the details of LSTM-based IONet. Each frame of inertial data is used as a input for single LSTM cell: $\mathbf{x}_i = (\mathbf{a}_i, \mathbf{w}_i)$. 
In the recurrent model, a hidden state $\mathbf{h}_i$ containing the history information of inputs, is maintained and updated at each step $i$ by:
    \begin{equation}
        \mathbf{h}_{i+1} = \text{LSTM}(\mathbf{h}_i, \mathbf{x}_i) .
    \end{equation}
This recurrent process compresses the high dimensional inertial sequence to a high-level compact motion description $\mathbf{h}_i \in \mathbb{R}^m$. $m$ is the number of hidden states. Finally, after recurrently processing all the data, the last hidden feature $\mathbf{h}_n$ contains the compressed information of the entire sequence for the motion transformation prediction. 

In IONet model, the polar vector $(\Delta l, \Delta \psi) \in \mathbb{R}^2$ (the displacement of location and heading), which proves to be observable from a sequence of inertial data, is learned by LSTMs. After LSTMs, a fully-connected layer then maps the hidden features into the target polar vector:
\begin{equation}
    (\Delta l, \Delta \psi)=\text{FC}(\mathbf{h}_n).
\end{equation}
Subsequently, in a sequence with timesteps between [0, n], the location $(L^x_n, L^y_n)$ at the $n$ th step is updated by
	\begin{equation}
     	\label{eq: location}
    	\left\{
    	\begin{aligned}
    		L^x_n=L^x_0+\Delta l \cos(\psi_0+\Delta \psi) \\
        	L^y_n=L^y_0+\Delta l \sin(\psi_0+\Delta \psi),
        \end{aligned}
       \right.
    \end{equation}
where $(L^x_0, L^y_0)$, and $\psi_0$ are the beginning location and heading of the sequence. .  

Instead of building explicit model to describe human motion, such as Weinberg's model \cite{Weinberg2002}, IONet is able to model motion dynamics and temporal dependencies of sensor data implicitly.  
Compared with PDR models, IONet is not restricted to the empirical step model, but is capable of regressing the average velocity anytime, i.e. the location transformation during any fixed period of time.

\subsection{Lightweight Inertial Odometry Neural Networks}

\begin{figure}
    	\centering
        \includegraphics[width=0.5\textwidth]{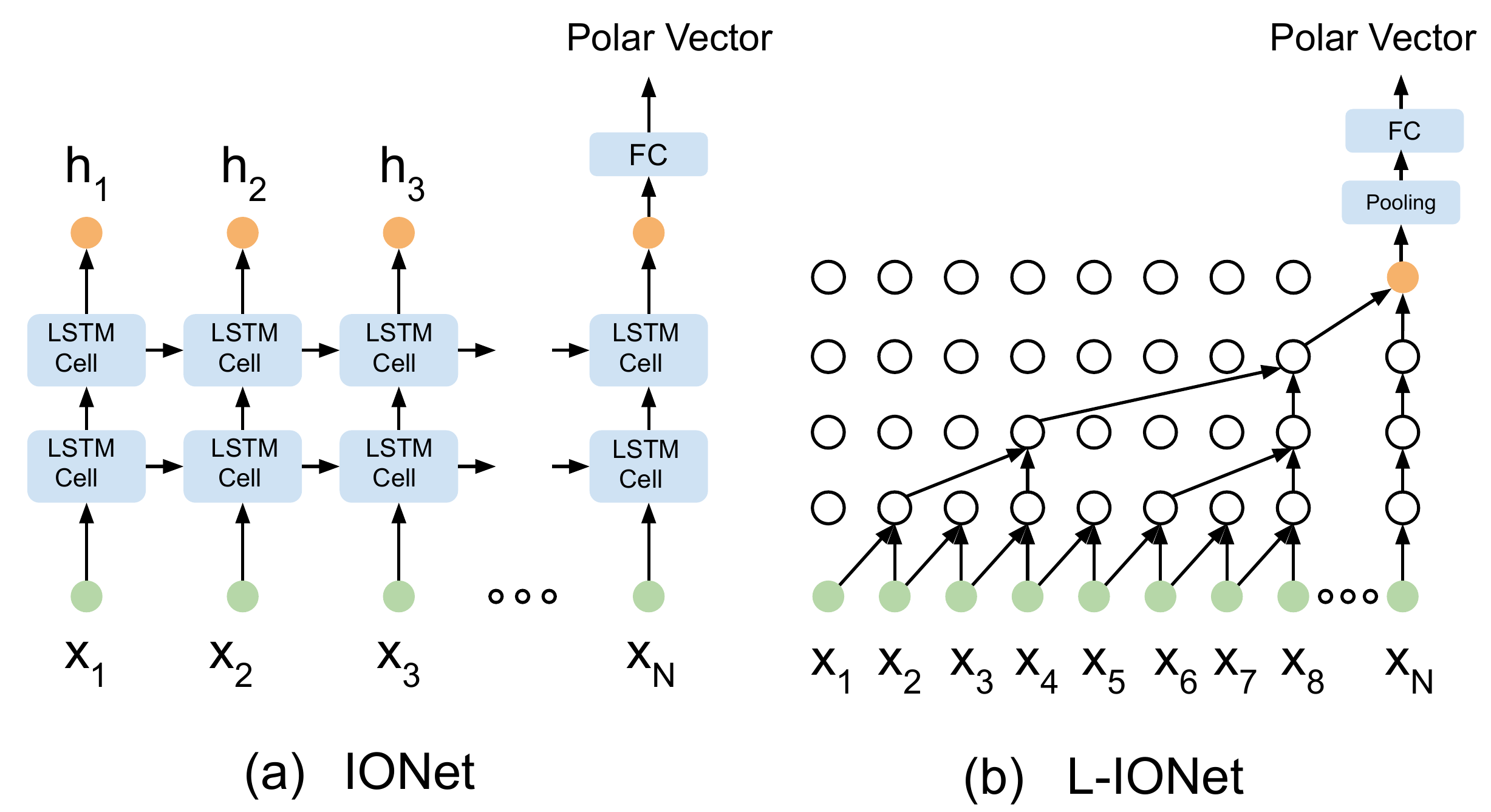}
        \caption{\label{fig:wavenet} A comparison of LSTM-based IONet and WaveNet-based L-IONet}
    \end{figure}

To this end, we introduce the Lightweight Inertial Odometry Neural Network (L-IONet), a lightweight framework to learn inertial tracking, which is more efficient in resource and computational consumption than the previous IONet approach. The detailed structure of L-IONet can be found in Figure \ref{fig:wavenet} (b).

Although deep learning solutions demonstrate great potential to solve sensing problems, e.g. IONet in inertial navigation, their huge computational and memory requirement slows down the deployment of DNN models onto low-end devices \cite{Yao2018}. 
Compared with deploying machine learning models on the cloud, computation at the edge reduces the bandwidth usage, cloud workload and latency.
Besides, it helps protect the users' privacy, as all sensor data hence remain at the users' device rather than uploading to the cloud \cite{Samie2019}. 
Therefore, it is necessary to design an efficient and fast DNN model to enable the inference of IONet on low-end devices, e.g. mobile phone, smartwatch, Raspberry Pi. 

The main bottleneck of IONet framework is the LSTM module. 
During the backpropagation of model training, recurrent models confront the so-called gradients vanishing problem \cite{greff2016lstm}, when processing long sequential data. This is especially the case in our inertial tracking task, as the input is a long sequence of 200 frames of inertial measurements, causing the optimization of recurrent models to be hard and unstable. Moreover, parallel training and inference is difficult for recurrent models, due to the fact that recurrent models have to exploit the sequential relation of the inputs and outputs. This limitation requires a sequence of input to be feed into recurrent models in order, consuming huge training time and computational resources to converge. In addition, the inference speed is a bottleneck for deploying RNN models on low-end devices, such as IoT devices or mobile phones, because of the complex operations inside recurrent networks. In contrast, the feed-forward models, e.g. WaveNet are more lightweight, and able to balance the trade-off between the accuracy and inference speed \cite{Oord2016}.

We propose to replace the recurrent model with an autoregressive model to produce outputs using the recent frames of a sequence. A good example is WaveNet, a generative causal autoregressive model, widely applied in processing speech and voice signals for synthesis tasks \cite{Oord2016}. Inspired by WaveNet, we propose an autoregressive model based L-IONet to process the long continuous signals of inertial sensors to predict polar vectors, which are further connected with previous states to reconstruct trajectories. Because L-IONet has no recurrent module and fewer complex nonlinear operations, this feed-forward model is easier to train in parallel, and much faster at state inference.

The basic module of our proposed framework is the causal dilated convolution layer. 
It can be viewed as a convolutional neural network (ConvNet) with a sliding window, but is a specific type of ConvNet that works perfectly on long sequential data. Compared with a regular ConvNet, the causal convolution inside our model is a 1-dimensional filter that convolves on the elements of current and previous timestep from last layer, to prevent using future states, as shown in Figure 4 (b). The stacked layers of dilated convolutions allow the receptive area of convolution operation to be made very large by using the convolution that skips input values with a certain distance. And hence the model is able to capture a long sequence of data without being too huge \cite{yu2015multi}.

Each causal dilated convolution performs via a gated activation unit:
    \begin{equation}
        \mathbf{z} = \text{tanh} (\mathbf{W}_{f,k}*\mathbf{x}) \odot \sigma (\mathbf{W}_{g,k}*\mathbf{x})
    \end{equation}
where $\mathbf{W}$ denotes the weights of the convolutional filters, $f$ and $g$ represent filters and gates, $k$ is the layer index, $*$ is the convolution operator, and $\odot$ is an element-wise multiplication operator. Meanwhile, residual and skip connection modules are adopted to enable a deeper structure, and improve the model's non-linearity in regression.

In our L-IONet framework, several layers of dilated causal convolutions are stacked to increase the receptive areas of the inputs. They skip the inputs with a specified stride, and their dilation doubled for every layer. In our case, an 8-layers model with a dilation of 1, 2, 4, 8, 16, 32, 64, 128 for each layer respectively, is enough to process a sequence of 200 frames of inertial data.  

The original WaveNet is designed for audio generation, which quantizes the real data into possible values, and reconstructs from the quantized data using the softmax function.
Instead of learning classification possibility, our framework replaces the softmax function with a pooling layer and a fully-connected layer to map the extracted features $\mathbf{z}$ into the 2-dimensional polar vectors: 
\begin{equation}
    (\Delta l, \Delta \psi)=\text{FC}(\text{Pool}(\mathbf{z})) .
\end{equation}
Similar to IONet model, the predicted polar vectors are further connected with previous system states to produce current locations via Equation \ref{eq: location}.

Compared with IONet, the main advantages of our proposed L-IONet can be summarized in two-folds -
1) Accuracy: the WaveNet module inside our L-IONet is extremely suitable to processing long continuous sensor signals, i.e. inertial data in our case. Compared with recurrent models, WaveNet mainly consists of dilated causal convolutions, which is easier to optimize, converge and recover optimal parameters during training, as no hidden states need to be updated and maintained for a long sequence of input data. Besides, the residual modules inside WaveNet further improves the expressive capacity of our model and thus provides more accurate polar vector predictions. 
2) Efficiency: L-IONet shows large improvement of both training and inference speed over IONet, enabling it to be deployed onto low-end devices easily, as illustrated in the experiments of Section V. On one hand, WaveNet allows parallel training. On the other hand, the convolutional operations in our L-IONet are faster to perform than the complex operations inside recurrent models \cite{Yao2018}.
The power of feed-forward models (e.g. WaveNet or Transformers \cite{vaswani2017attention}) have already shown huge improvement in voice synthesis and machine translation \cite{devlin2019bert}. Recently, there is a trend to replace the LSTM module with feed-forward models in sequence modelling tasks. However, few work has explored the potential applications of feed-forward models in processing continuous sensor data, e.g. inertial data as inertial tracking in our case.

\subsection{Sliding Window}

 \begin{figure*}
    	\centering
    	\begin{subfigure}[t]{0.32\textwidth}
        	\includegraphics[width=\textwidth]{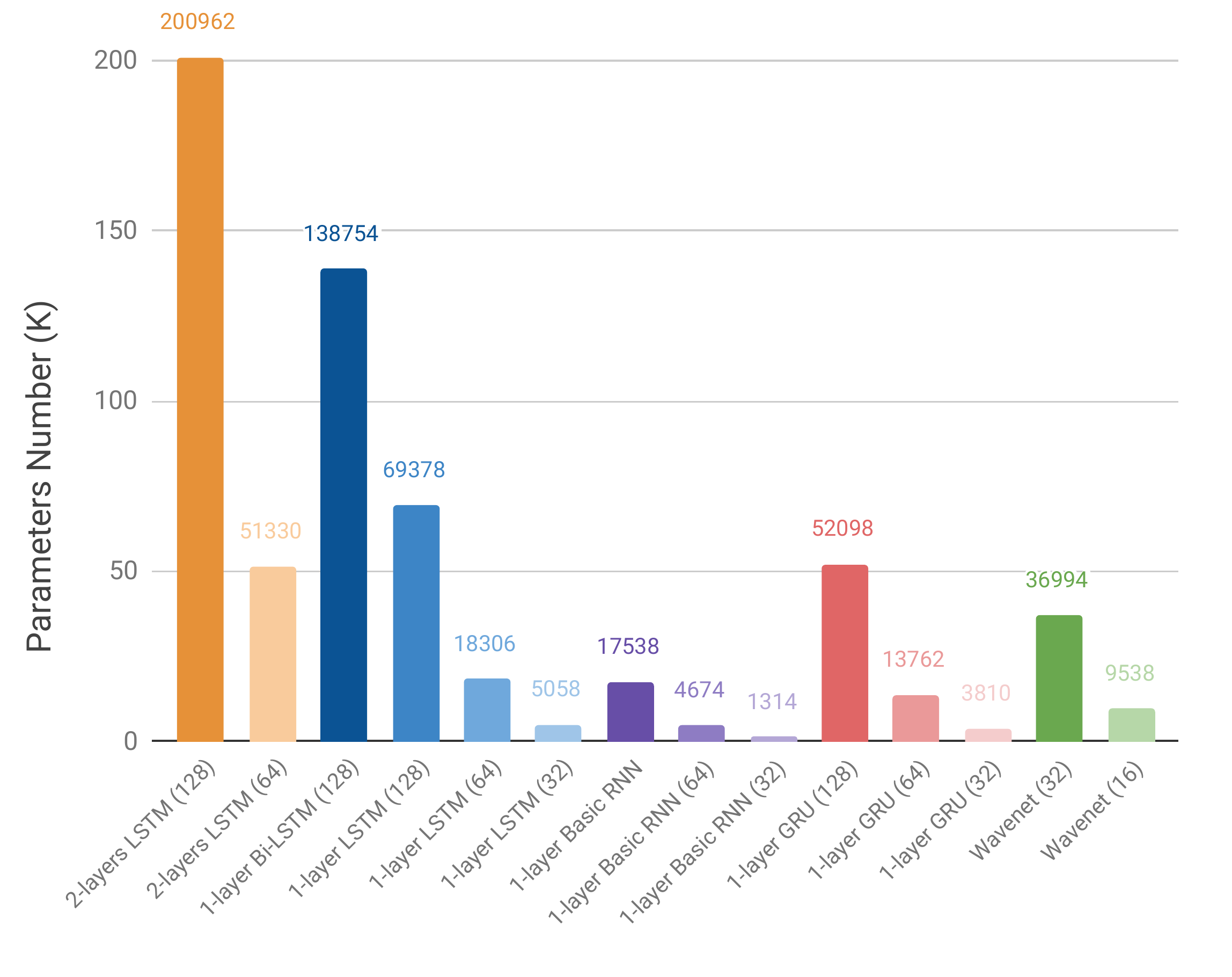}
        	\caption{\label{fig:parameters} Parameters Number}
        \end{subfigure}
        \begin{subfigure}[t]{0.32\textwidth}
        	\includegraphics[width=\textwidth]{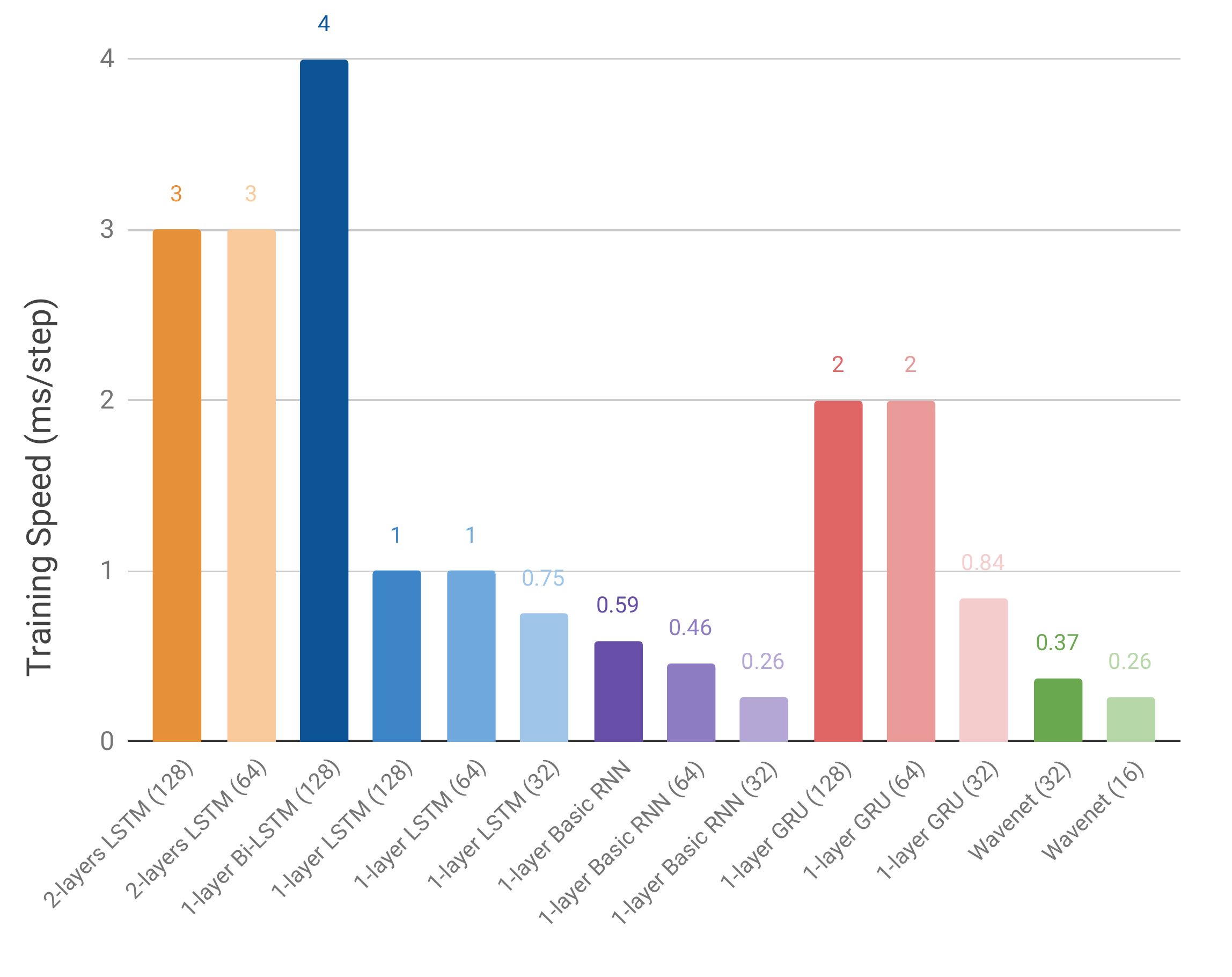}
        	\caption{\label{fig:training speed} Training Speed}
        \end{subfigure}
        \begin{subfigure}[t]{0.32\textwidth}
        	\includegraphics[width=\textwidth]{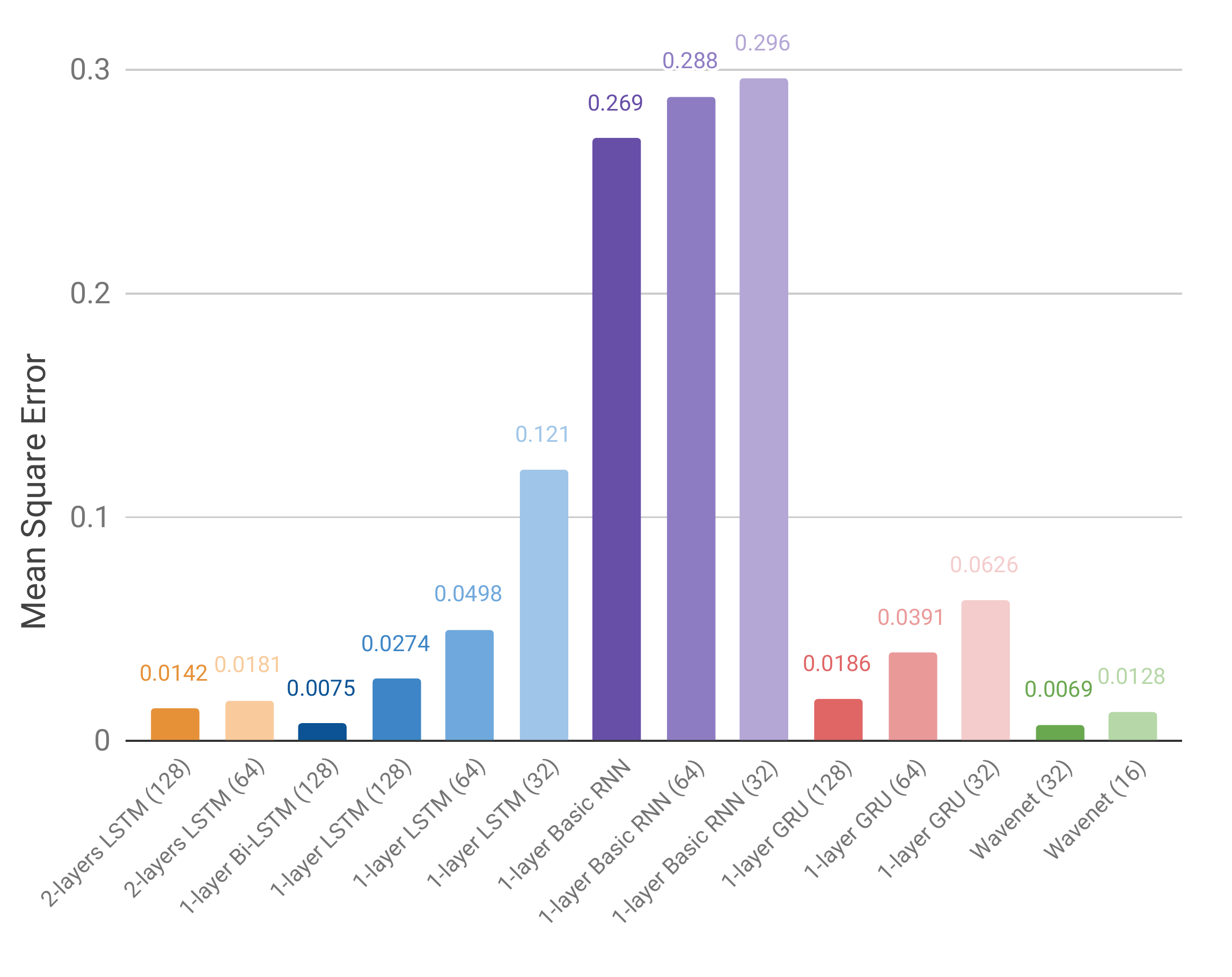}
        	\caption{\label{fig:accuracy} Mean square error}
        \end{subfigure}
        \caption{\label{fig:edge} A comparison of IONet and L-IONet models in terms of their (a) number of parameters, (b) training (convergence) speed and (c) test accuracy. L-IONet shows competitive performance to IONet, but requires less memory and a quicker training time.}
    \end{figure*}
    
       \begin{table*}[ht]
        \caption{The execution time (ms) of the deep neural networks models on the low-end devices. }
        \label{tb: inference time}
        \begin{center}
        \begin{tabular}{c c c c c}
        \hline
        Models & Huawei Mate 8 & Nexus 5 & HTC One M8 & Sony SW2\\
        \hline
        2-layers LSTM (128) & 38.13 & 38.65 & 88.13 & 342.61\\
        2-layers LSTM (64) & 11.42 & 14.74 & 33.62 & 109.41\\
        1-layer Bi-LSTM (128) & 27.23 & 31.08 & 71.02 & 261.08\\
        1-layer LSTM (128) & 12.7 & 16.15 & 37.38 & 130.85\\
        1-layer LSTM (64) & 4.65 & 7.08 & 16.69 & 48.9\\
        1-layer LSTM (32) & 1.32 & 2.25 & 3.49 & 18.7\\
        1-layer Basic RNN (128) & 2.4 & 3.13 & 4.63 & 31.2\\
        1-layer Basic RNN (64) & 0.86 & 1.7 & 2.69 & 14.06\\
        1-layer Basic RNN (32) & 0.46 & 1.13 & 1.94 & 8.78\\
        1-layer GRU (128) & 7.29 & 12.92 & 14.72 & 81.8\\
        1-layer GRU (64) & 3.02 & 6.21 & 8.00 & 35.03\\
        1-layer GRU (32) & 1.76 & 4.24 & 5.68 & 21.54\\
        \hline
        WaveNet (32) & 3.7 & 6.47 & 13.74 & 56.78\\
        WaveNet (16) & 1.27 & 3.58 & 8.43 & 27\\
        \hline
        \end{tabular}
        \end{center}
    \end{table*}

In order to increase the output rate of neural network predictions, we present a sliding window method.
As Figure \ref{fig:framework} (b) and (c) demonstrated, the inertial sensory readings are segmented into independent sequences by using a fixed-size sliding window.
In our problem, we choose $n$ the window size of the sequence as 200 frames (2 seconds), with a stride for sliding the window as 10.
The polar vector is predicted by the deep neural networks from each sequence, and connected by a merging module to generate locations, as Equation (3) described. Note that the current location is updated with the location 200 frames before it rather than the previous states 10 frames before it. With the predictions from the overlapping windows, the output rate is increased onto 10 Hz.  
Low pass filters are further used upon the polar vectors and locations to smooth the predictions for trajectory reconstruction.

\section{Experiments}
In this section, we implemented and trained the IONet and L-IONet models on the OxIOD dataset, and conducted extensive experiments to evaluate their performance on the low-end devices, velocity estimation, and localisation experiments. 

\subsection{Setup}
\textbf{Training and Testing:}
We split the dataset into the training and testing set for each attachment scenario, i.e. handheld, pocket, handbag and trolley. The detailed description can be found in the dataset folder. All the data is split using a window size of 200 and a stride of 10. Considering the convenience of deploying on devices, our IONet and L-IONet were implemented in the Keras framework with a Tensorflow backend. By minimising the mean square error between the estimated values and ground truth provided by our dataset, the optimal parameters were trained via the ADAM optimiser \cite{Kingma2014} with a learning rate of $1e^{-5}$. The batchsize is chosen as 256. We trained each of the model configurations on one NVIDIA TESLA K80.

\textbf{Devices:}
To evaluate the performance of our proposed models on low-end devices, we chose three levels off-the-shelf consumer smartphones, i.e. Huawei Mate 8, Nexus 6, HTC One M8, and one consumer smartwatch, i.e. Sony Smartwatch 2. Huawei Mate 8 is equipped with octa-core (4x2.3 GHz and 4x1.8 GHz) CPU and 4 GB RAM. Nexus 6 is equipped with quad-core 2.7 GHz CPU and 3 GB RAM. HTC One M8 is equipped with quad-core 2.5 GHz CPU and 2 GB RAM. Sony Smartwatch 2 is equipped with 1 core 180 MHz CPU and 256 MB RAM. Our IONet and L-IONet models were first trained with the Keras framework on GPUs, further converted into Tensorflow Lite models, and then deployed on the low-end devices to test their inference speed.

\subsection{Model Performance at the edge}

We conducted a systematic research into the inference performance of DNNs models for inertial tracking at the edge. 
The LSTM-based IONet is compared with our proposed WaveNet style L-IONet, with different hyperparameters chosen to demonstrate their impacts on the model performance, which are the number of layers, whether LSTMs are bi-directional or not, the number of hidden states for LSTMs, and the number of convolutional filters for WaveNet. Moreover, we replaced the LSTM module in IONet with
GRUs and Basic RNNs as baselines to show the trade-off between model accuracy and efficiency.

Figure \ref{fig:edge} compares different model configurations of IONet (LSTM), IONet (GRU), IONet (Basic RNN) and L-IONet (WaveNet), in terms of their number of parameters, training speed and mean square error (MSE) of predicted polar vectors. It is clear to see that the L-IONet with 32 filters i.e. WaveNet (32), achieves the highest accuracy, with a prediction error of 0.0069, even slightly lower than that of IONet with 1-layer Bi\_LSTM (128), i.e. 1-layer Birectional LSTM with 128 hidden states. In contrast, the number of parameters in the L-IONet with WaveNet (32) is only one quarter that of the IONet with 1-layer Bi-LSTM (128). Meanwhile, L-IONet with WaveNet (32) is around 10 times faster than IONet with 1-layer Bi-LSTM when training on a Tesla K80 GPU. 
This indicates that L-IONet shows competitive performance in accuracy over IONet, while still superior in the speed and resource consumption.

Table \ref{tb: inference time} illustrates the execution time of different IONet (LSTM, GRU, Basic RNN) and L-IONet (WaveNet) models when deployed on Huawei Mate 8, Nexus 5, HTC One M8 and Sony SW2 respectively. The execution time (milliseconds, ms) is the average inference time of these models at the low-end devices. 
The L-IONet models, i.e. WaveNet (32) and WaveNet (16) performed faster inference than the LSTM-based IONet models. Even at the swartwatch device equipped with very limited CPU and memory, our proposed L-IONet is capable of realising real-time inference, producing outputs within only 56.78 ms (WaveNet (32)) and 27 ms (WaveNet (16)) for each step. The inference speed of IONet with 1-layer LSTM (64) is similar to WaveNet (32), but its prediction error is almost 8 times higher than WaveNet (32).  
We further compare LSTM (32) with WaveNet (32) and find that the prediction error of LSTM (32) increases to 17.5 times higher than WaveNet (32), although LSTM (32) is with faster inference speed.   
It is interesting to see that GRUs are more lightweight compared with both LSTM and WaveNet models. However, the prediction accuracy of GRU models are not satisfying with larger prediction errors, i.e. 0.0186, 0.0391, and 0.0626 for GRU(128), GRU(64) and GRU (32) respectively, around 3, 6, 9 times higher than WaveNet (32). The Basic RNNs (128) (64) (32) have fewer parameters, and faster inference speed on low-end devices than our WaveNet-based L-IONet, but they almost learned nothing from inertial data with huge test errors (i.e. 0.268, 0.288 and 0.296), nearly 40 times larger than the WaveNet models. Therefore, our WaveNet based L-IONet models show better trade-off between the prediction accuracy and on-device inference efficiency. 
It is worth noticing that the Wavenet-based L-IONet owns the advantages of faster training speed over Basic RNNs, LSTMs, and GRUs, as shown in Figure 5 (b). This is because feed-forward models are easier to train and optimize than recurrent models.

\subsection{Velocity and Heading Estimation}
\label{sec: velocity}

	\begin{figure*}
    	\centering
        \begin{subfigure}[t]{0.48\textwidth}
        	\includegraphics[width=\textwidth]{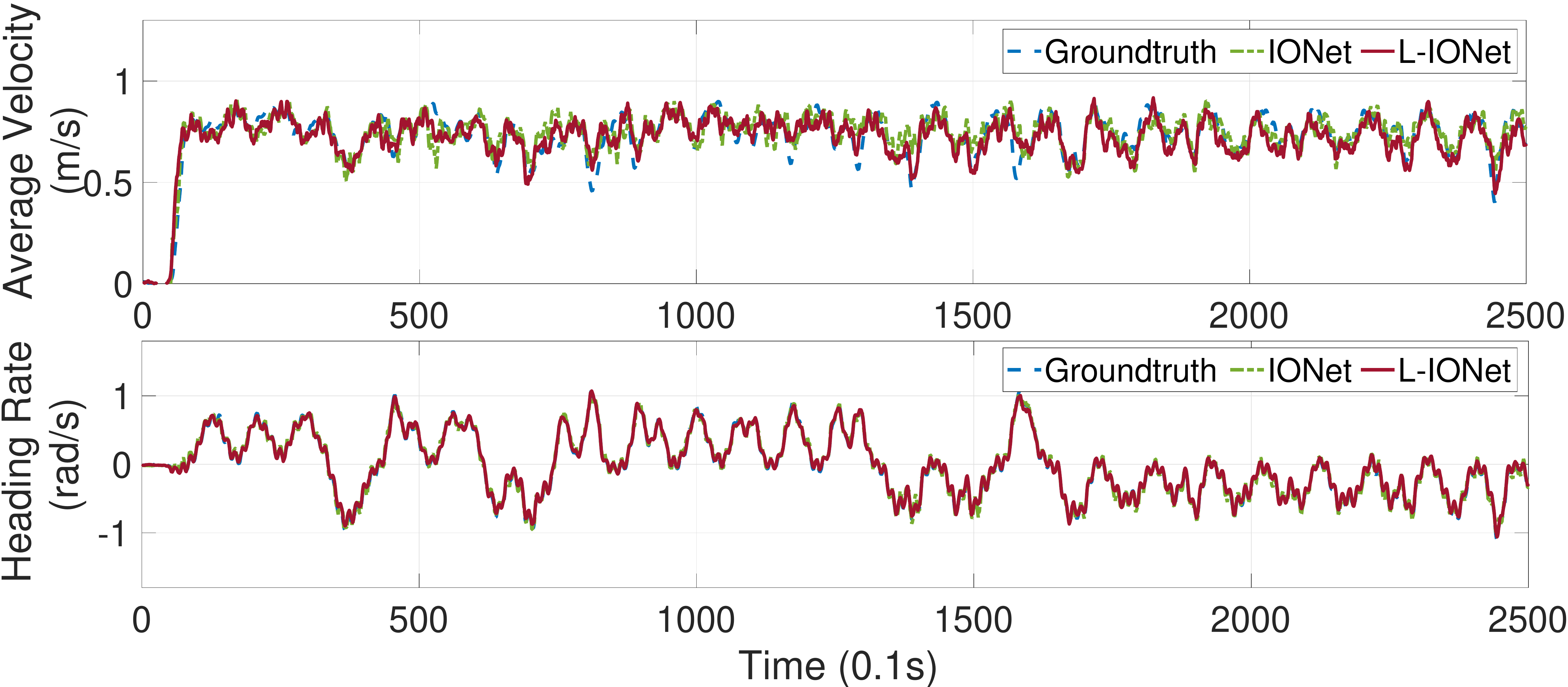}
        	\caption{\label{fig:speed_normal} Walking Normally}
        \end{subfigure}
        \begin{subfigure}[t]{0.48\textwidth}
        	\includegraphics[width=\textwidth]{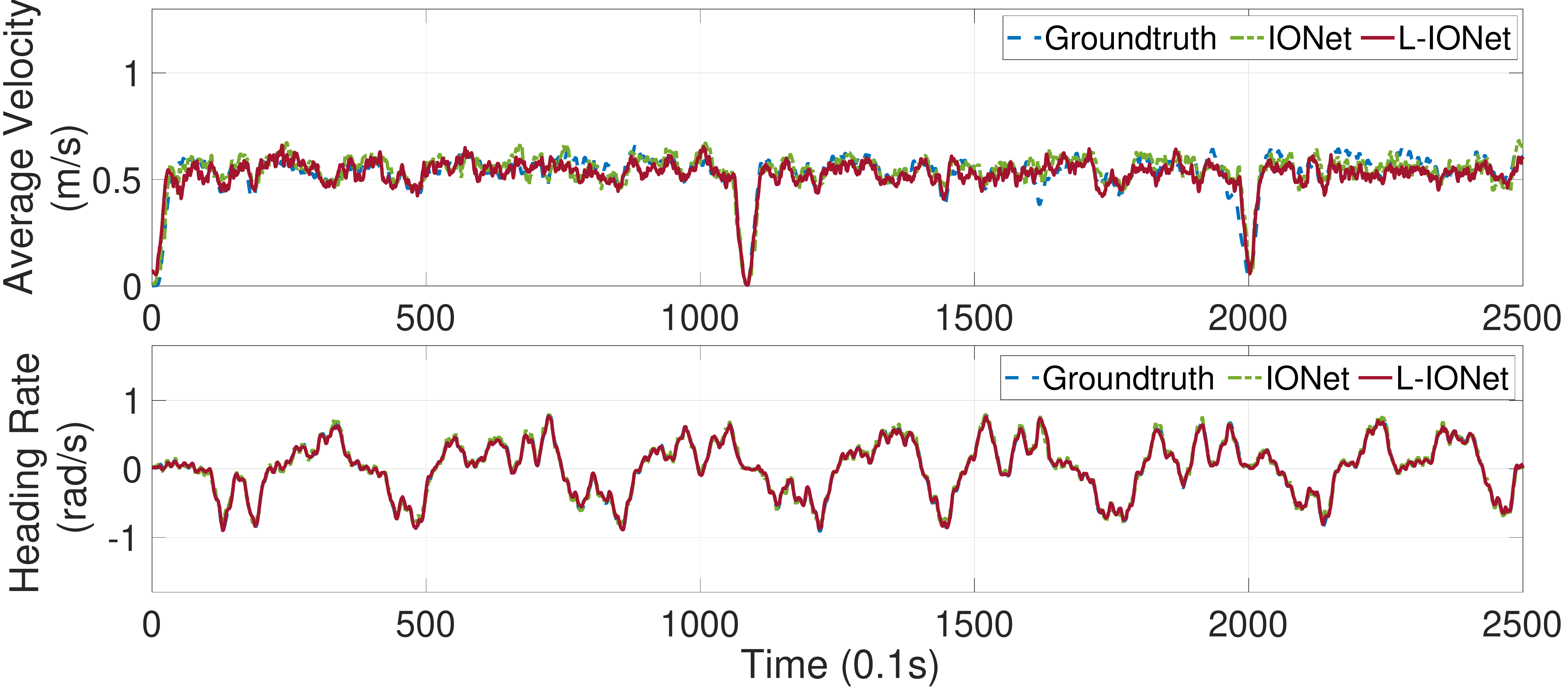}
        	\caption{\label{fig:speed_slow} Walking Slowly}
        \end{subfigure}
        \begin{subfigure}[t]{0.48\textwidth}
        	\includegraphics[width=\textwidth]{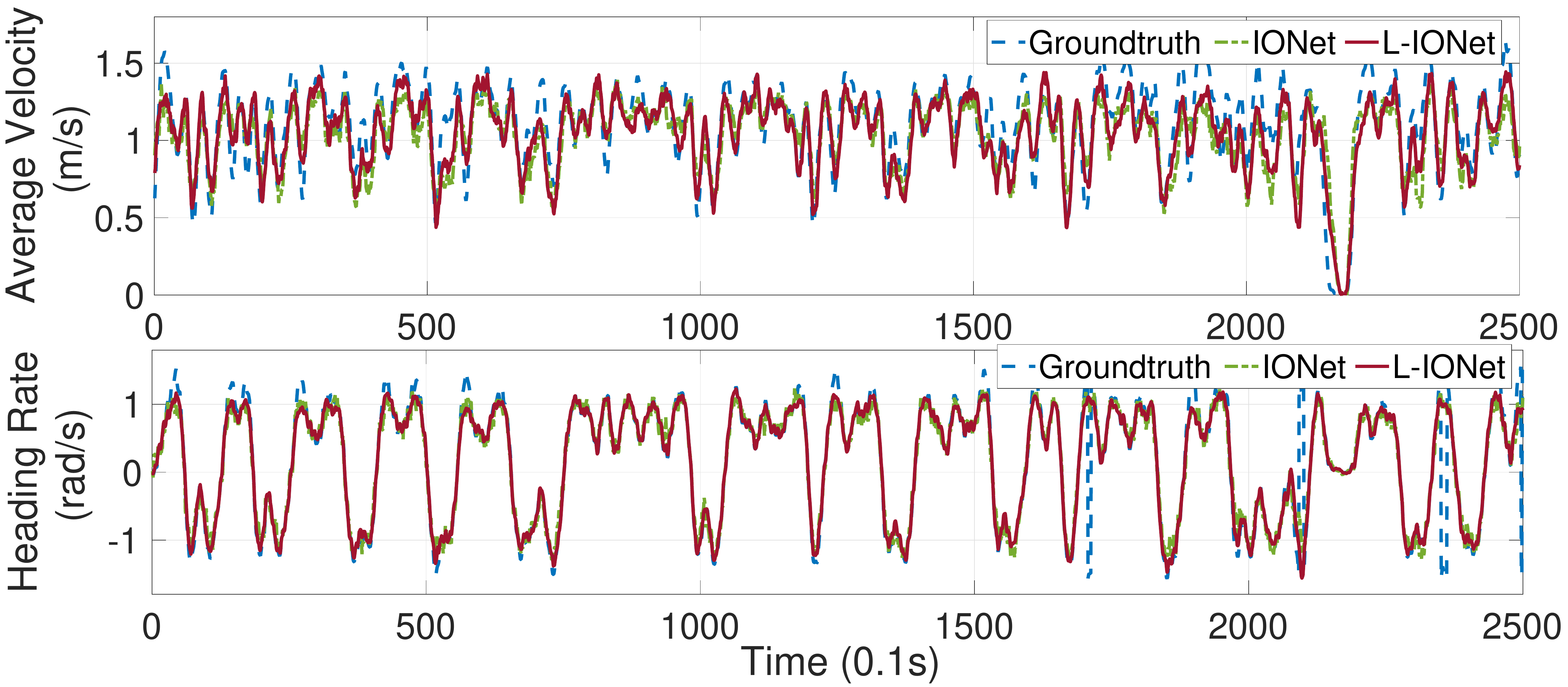}
        	\caption{\label{fig:speed_run} Running}
        \end{subfigure}
        \begin{subfigure}[t]{0.48\textwidth}
        	\includegraphics[width=\textwidth]{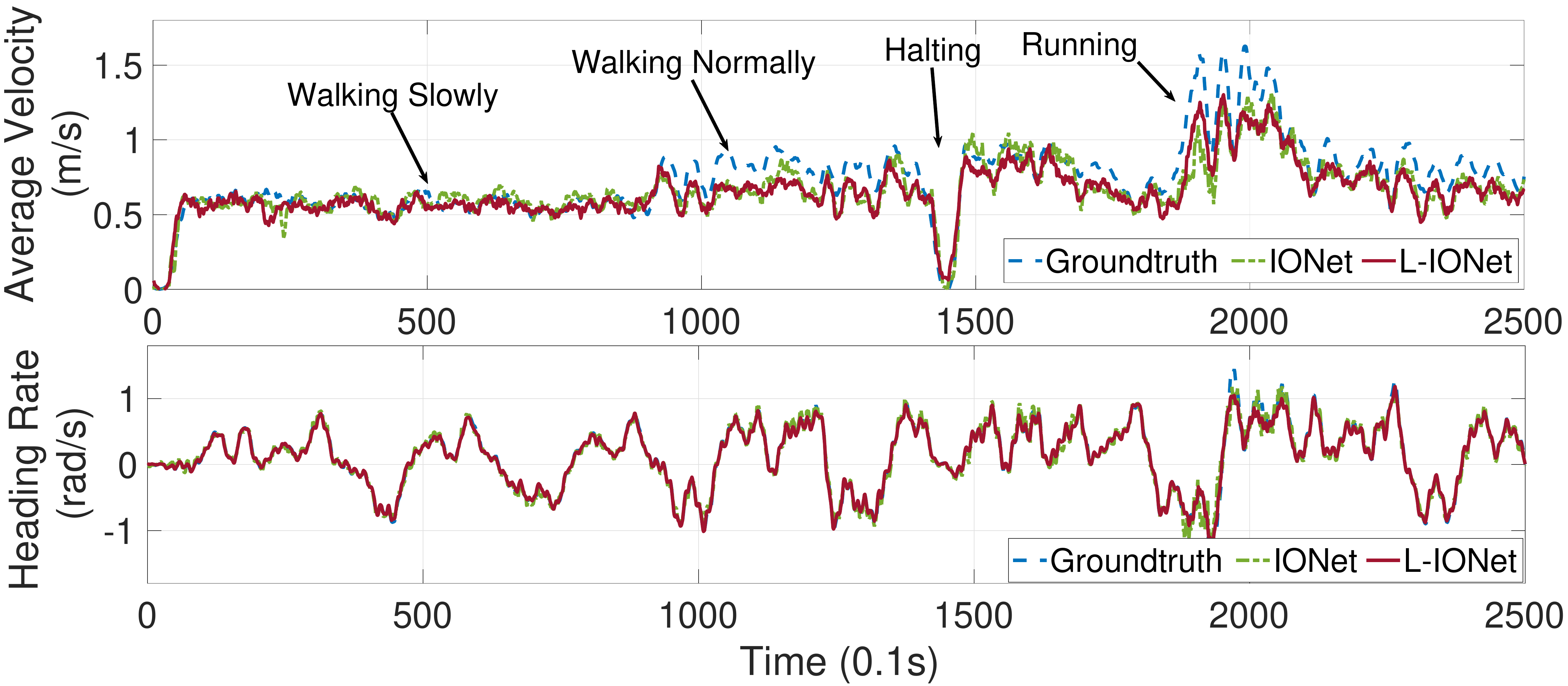}
        	\caption{\label{fig:speed_mix} Mixed Activities}
        \end{subfigure}
        \caption{\label{fig:velocity estimation} The velocity and heading estimations for a) walking normally, b) walking slowly, c) running and d) mixed motion modes. The ground truth was captured by Vicon System, while the values from IONet and L-IONet were predicted by the learning model trained on our proposed dataset.}
    \end{figure*}

As a demonstration of training performance, the IONet and L-IONet models were trained on the OxIOD dataset to predict the average velocity and heading rate of pedestrian motion.
The average velocity $\bar{v}$ and heading rate $\dot{\psi}$ are defined as the location displacement $\Delta l$ and heading change $\Delta \psi$ during a window size of time $n$:
	\begin{equation}
		(\bar{v}, \dot{\psi}) = (\Delta l / n, \Delta \psi / n) .
	\end{equation}
In our experiment setup, the window size $n$ was chosen as 2 seconds, so a sequence of inertial data (200 frames) $(\{(\mathbf{a}_i, \mathbf{w}_i)\}_{i=1}^{n})$ is fed into IONet or L-IONet model to predict the average velocity $\bar{v}$ and heading rate $\dot{\psi}$:
	\begin{equation}
		(\bar{v}, \dot{\psi}) = \text{IONet or L-IONet} (\{(\mathbf{a}_i, \mathbf{w}_i)\}_{i=1}^{n}) ,
	\end{equation}
where we used IONet with 1-layer 128-dimensional Bidirectional LSTM, and L-IONet with 32-filters WaveNet for training and prediction.

The training data are from the training sets of three motion modes categories: walking normally (handheld, 20 seqs), walking slowly (7 seqs) and running (6 seqs). To test its generalisation ability, we performed randomly walking in the Vicon Room, and used the trained neural network to predict the values for selected three motion modes and a mix of activities respectively. 
Fig. \ref{fig:velocity estimation} indicates that both IONet and L-IONet can model a variety of complex motions, and generalise well to mixed activities. The more lightweight L-IONet does not suffer an accuracy loss in this task.

\subsection{Deep Learning based Pedestrian Inertial Navigation}

We show how to solve the pedestrian inertial navigation problem using deep neural networks with the aid of our proposed OxIOD dataset. IONet and L-IONet models can reconstruct trajectories from raw IMU data, and provide users with their accurate locations. A 1-layer Bidirectional LSTM with 128 dimensional hidden states was adopted for IONet, while the WaveNet with 32 filters was used in L-IONet for the evaluation. Both models were trained with the above detailed split training sets from the four attachment categories, i.e. the handheld (20 sequences), pocket (10 sequences), handbag (7 sequences) and trolley (12 sequences). 
Two sets of experiments were conducted to evaluate our proposed models.

 \begin{figure*}
    	\centering
        \begin{subfigure}[t]{0.48\textwidth}
        	\includegraphics[width=\textwidth]{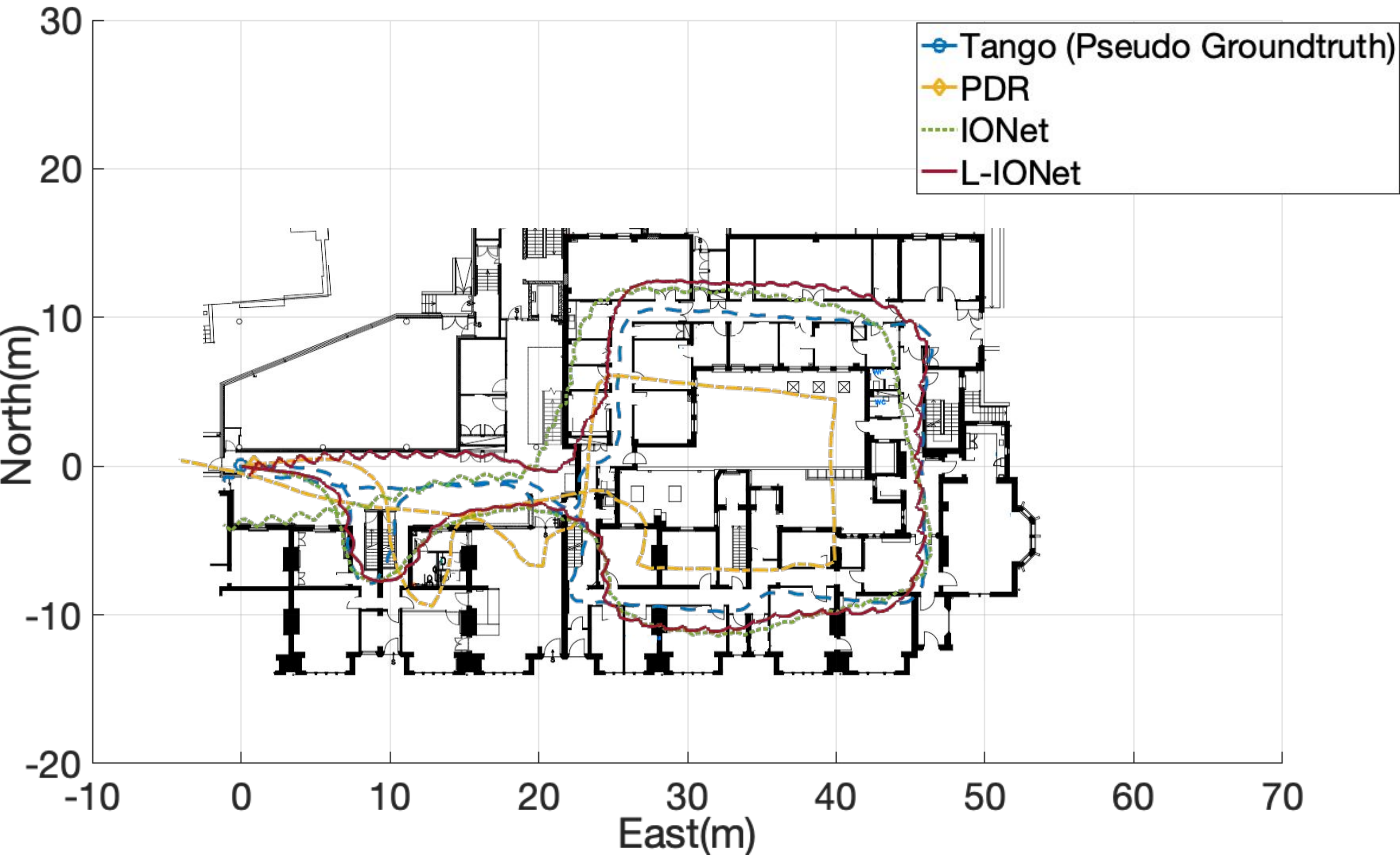}
        	\caption{\label{fig:floor1} Office Floor 1}
        \end{subfigure}
        \begin{subfigure}[t]{0.48\textwidth}
        	\includegraphics[width=\textwidth]{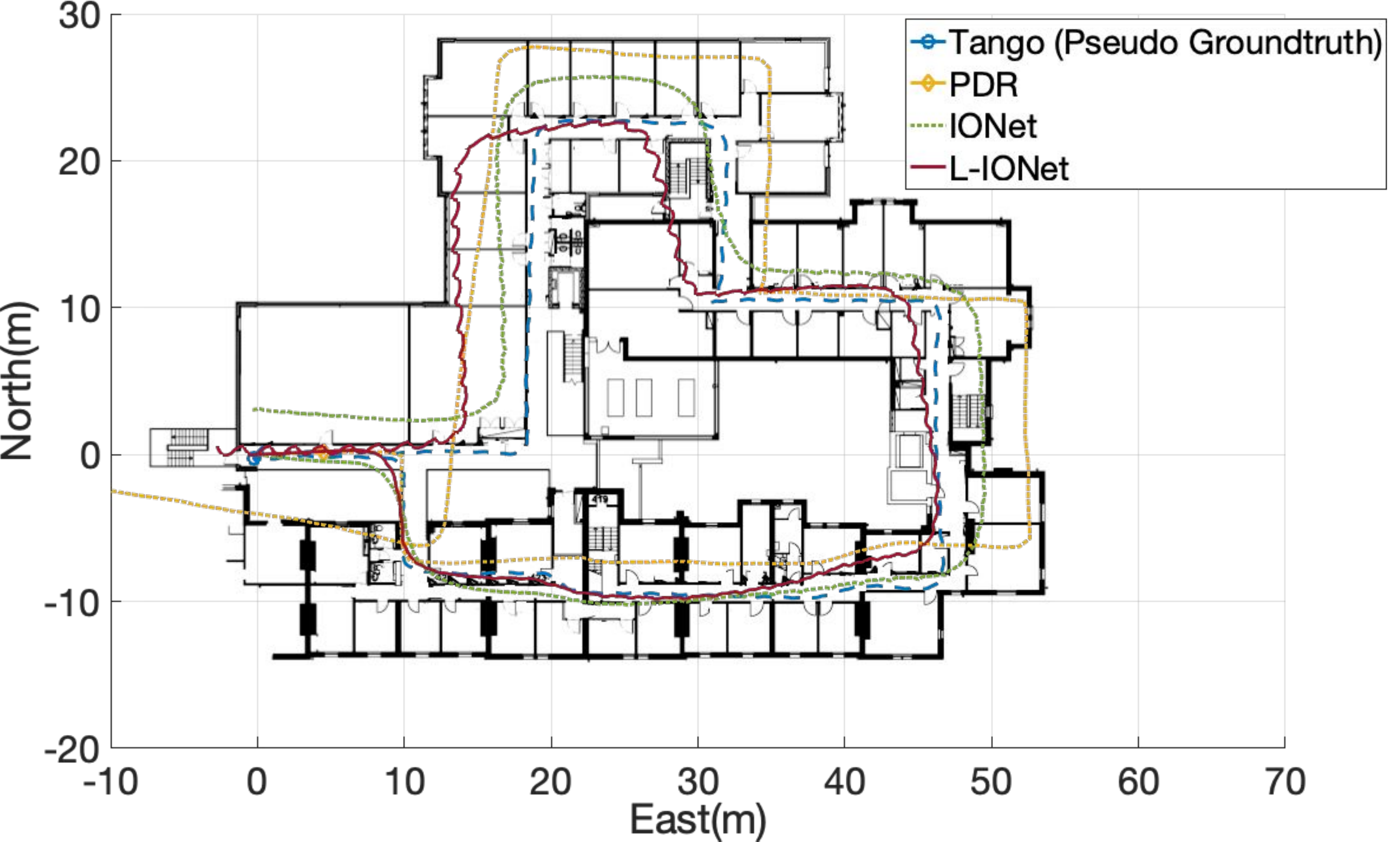}
        	\caption{\label{fig:floor2} Office Floor 2}
        \end{subfigure}
        \caption{\label{fig: largescale} The largescale localisation experiments were conducted on (a) Office Floor 1 and (b) Office Floor 2. The trajectories were generated from the IONet, L-IONet and PDR. The pseudo ground truth was provided by Google Tango device.}
    \end{figure*}

	\begin{figure*}
    	\centering
        \begin{subfigure}[t]{0.24\textwidth}
        	\includegraphics[width=\textwidth]{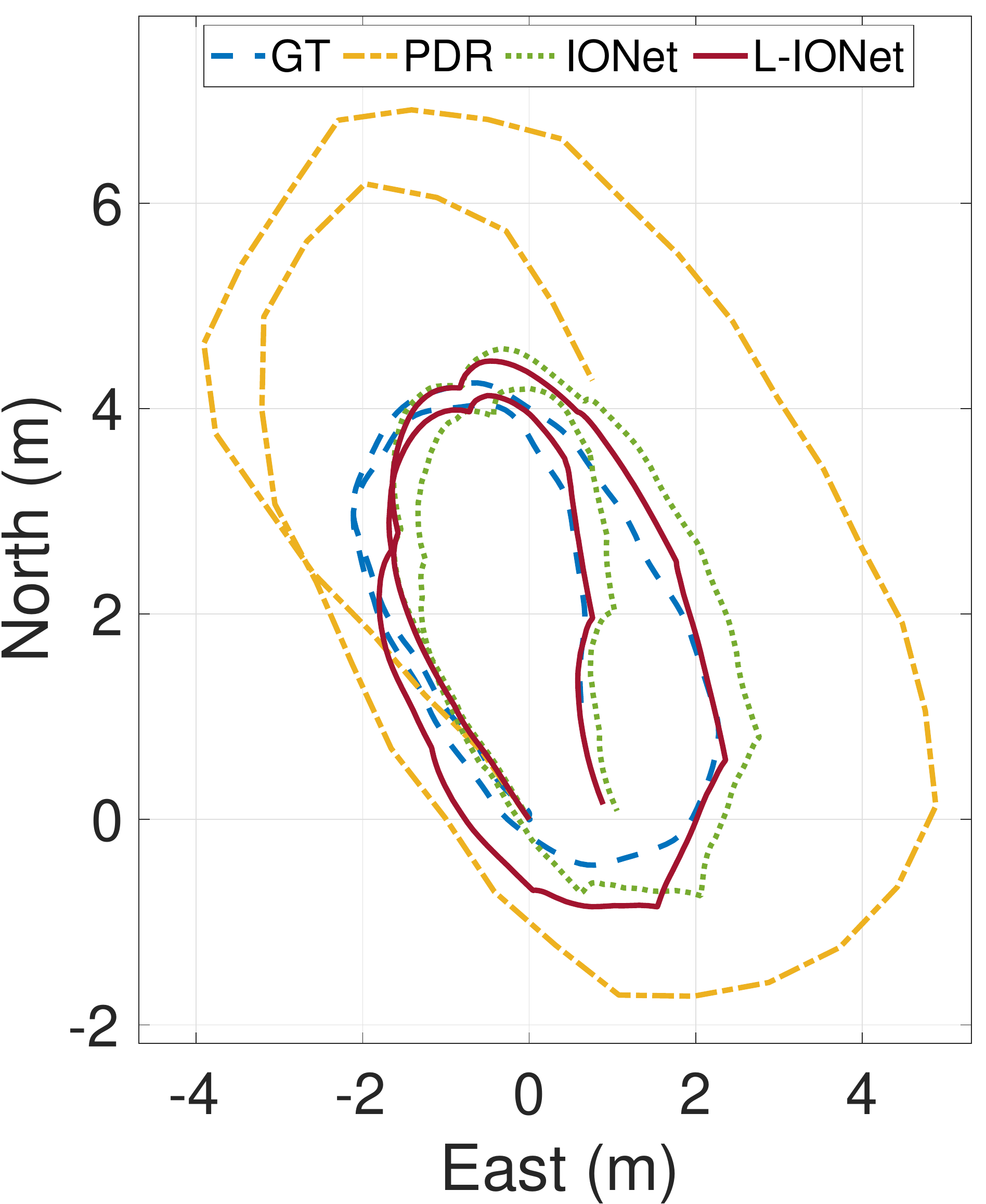}
        	\caption{\label{fig:handheld} Handheld}
        \end{subfigure}
        \begin{subfigure}[t]{0.24\textwidth}
        	\includegraphics[width=\textwidth]{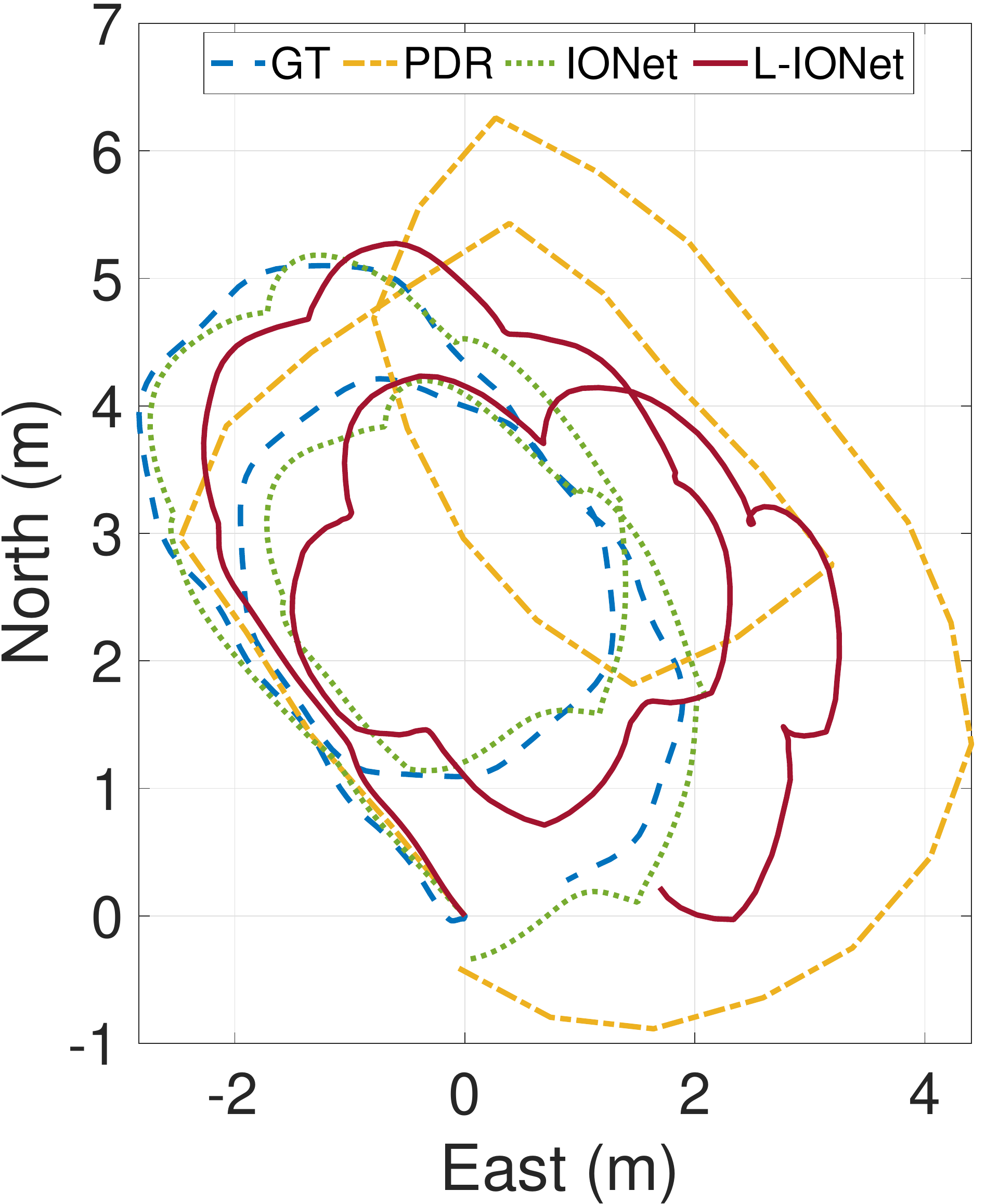}
        	\caption{\label{fig:pocket} In Pocket}
        \end{subfigure}
        \begin{subfigure}[t]{0.24\textwidth}
        	\includegraphics[width=\textwidth]{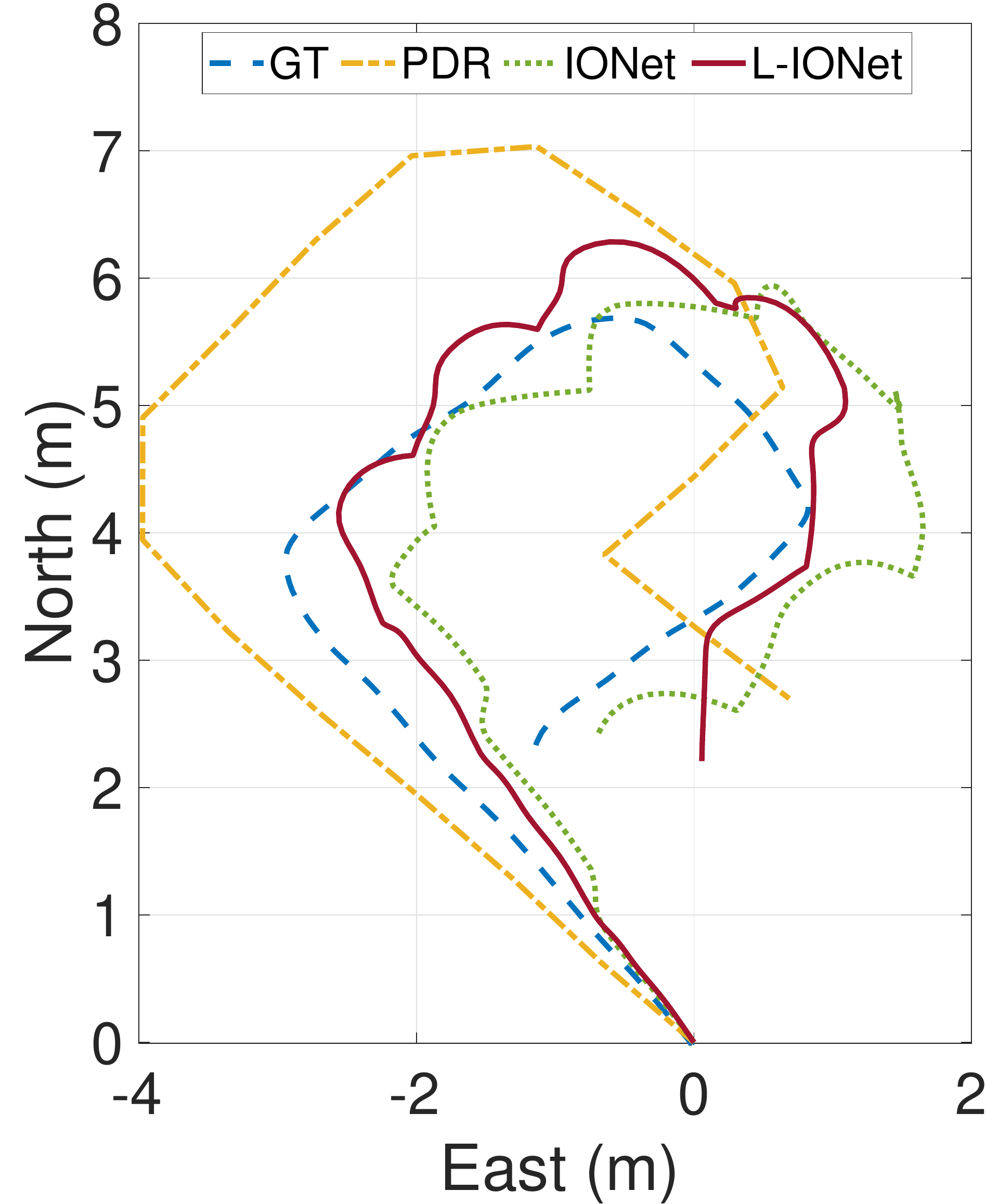}
        	\caption{\label{fig:handbag} In Handbag}
        \end{subfigure}
        \begin{subfigure}[t]{0.24\textwidth}
        	\includegraphics[width=\textwidth]{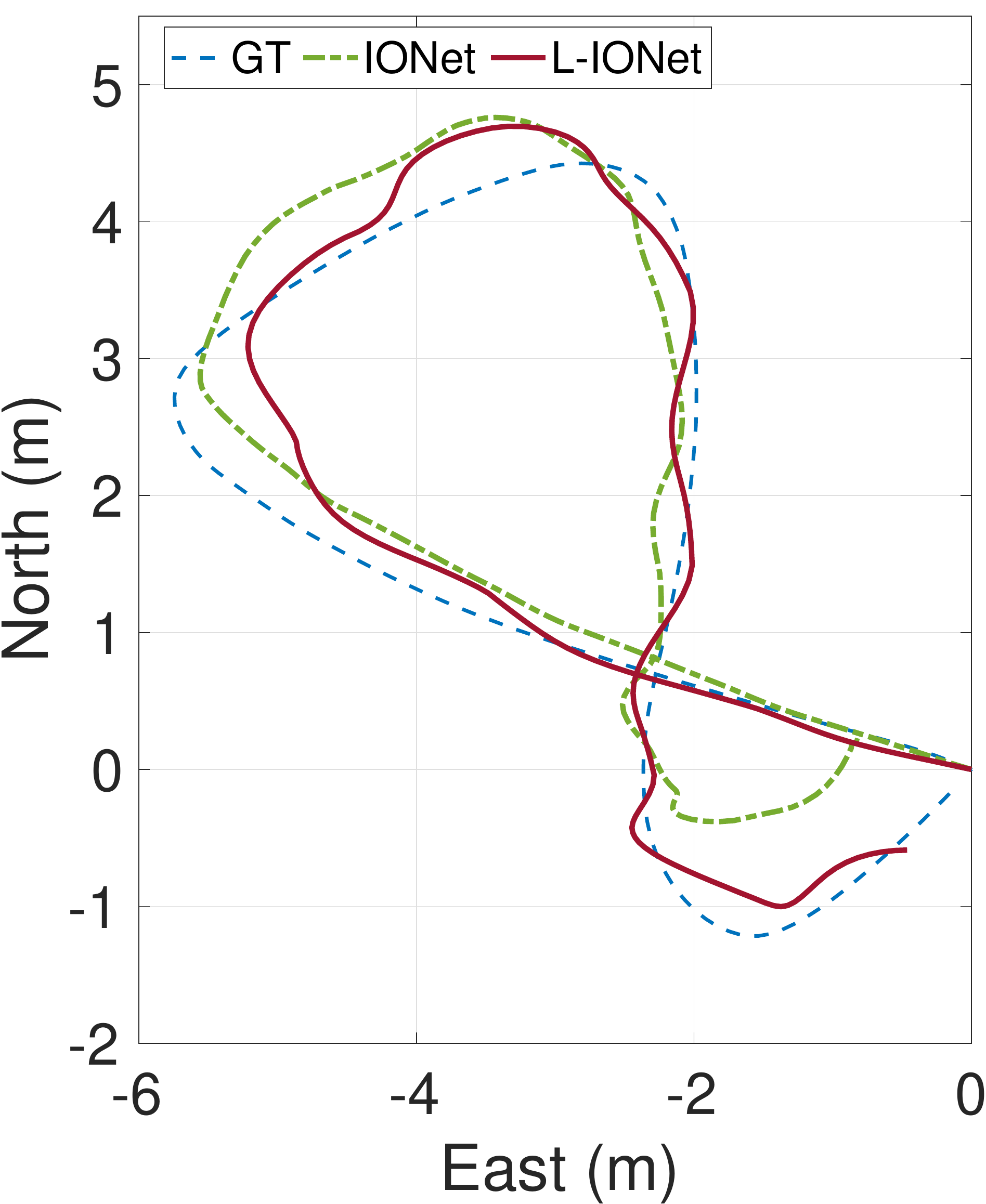}
        	\caption{\label{fig:trolley} On Trolley}
        \end{subfigure}
        \caption{\label{fig:deep tracking} The trajectories reconstruction for pedestrian tracking with device in four attachments: a) in the hand, b) in the pocket, c)in the handbag, and d) on the trolley respectively. The trajectories were generated from IONet, L-IONet and a basic PDR algorithm. PDRs do not work when the device was placed on the trolley, as no step can be detected in this situation. The ground truth values are provided by the Vicon System.}
    \end{figure*}

The first set of tests involved tracking a pedestrian with the phone in different attachments. In this experiment, the participant carrying the smartphone was asked to walk normally inside the Vicon room. The IMU data \footnote{The test data can be found at the 'test' fold of our dataset} were collected and feed into the IONet and L-IONet to predict the participant's motion. The installed motion capture systems can provide highly precise trajectories as groundtruth. Note that these walking trajectories are not present in the training dataset. A basic PDR algorithm was implemented as a baseline, and we show that our dataset can also be used as a benchmark for conventional PDR algorithms. Figure \ref{fig:deep tracking} demonstrates the trajectories generated from the groundtruth (blue line), PDR (green line), IONet (orange line) and L-IONet (red line). It indicates that the deep learning based methods outperformed the model-based PDR when the phone was placed either in the hand, pocket or handbag. The trolley tracking is a difficult problem for PDR algorithms, as no step (periodicity pattern) can be detected in this case, and hence a handcrafted model is hard to build for this wheeled motion. In contrast, the learning based approaches are still able to generalise to this general motion, and reconstruct physically meaningful trajectories, while the PDR algorithm fails in this task. The L-IONet model produced results even closer to the groundtruth compared with IONet, especially in the handheld and trolley domains. 
Figure \ref{fig:pocket} indicates that the IONet shows better performance when mobile device is inside pocket. This is because inertial sensor experiences less motion-dynamics inside pocket (pocket can be viewed to constrain mobile device when users are moving), so that LSTM based IONet is already good enough to capture its self-motion from inertial data. In the other domains, the WaveNet based L-IONet is more capable of representing the free motion of sensor, due to the expressive capability of WaveNet in processing long sequential data.

The other set of experiments is to perform large-scale localisation on two floors of an office building. Although DNN models were trained with inertial data collected inside the Vicon room, we show that the models can be used to predict the pedestrian motion outside the room directly. This is due to the fact that inertial data are not sensible to environments, and hence the proposed DNN models can generalize to new environment easily. Figure \ref{fig: largescale} demonstrates that both IONet and L-IONet models achieve good localisation results, although the two models never saw any data outside the Vicon room. This experiment shows the generalisation ability of the deep learning based models towards new environments.

\section{Conclusions and Future Work}
In this work, we propose L-IONet, a lightweight deep neural networks framework to learn inertial tracking from raw IMU data. L-IONet shows competitive performance over previous deep inertial odometry models. Meanwhile, L-IONet is more efficient in memory, inference and training. We conducted a systematic research into the performance of deep learning based inertial odometry models on low-end devices. Our L-IONet is able to achieve real-time inference on different levels smartphones, and even the smartwatch with very limited computational resources. Moreover, we present and release OxIOD, an inertial odometry dataset for training and evaluating inertial navigation models. 
With the release of this large-scale diverse dataset, it is our hope that it will prove valuable to the community and enable future research in long-term ubiquitous ego-motion estimation. 

Future work would include collecting data from more challenging situations, for example, 3D tracking. We plan to create on-line common benchmark and tools for the comparison of odometry models.
We also hope to include more IoT devices into our dataset, such as smartwatch, wristband, smart earphones, to extend the potential applications of this research.
A further extension to current deep inertial odometry models is to adopt knowledge distillation to compress the deep neural networks, which can reduce the number of parameters and enable faster training and on-device inference. 
Another future research direction is to investigate how to formulate dilated casual convolutional model (i.e. WaveNet style model) as a generic framewor to process a variety of sensor data e.g. temperature, pressure, light intensity, magnetic field, in other potential IoT application domains, e.g. health/activity monitoring, sport analysis, smart home and intelligent transportation.
Except the deep neural networks discussed above, other machine learning models might also be applied into data-driven IoT research domains, for example, Deep Belief Network (DBN) or Broad Learning System (BLS).

\bibliographystyle{IEEEtran}
\bibliography{refs.bib}

\end{document}